\documentclass[fleqno,10pt]{olplainarticle}

\usepackage[]{hyperref}

\title{LitS: A novel Neighborhood Descriptor for Point Clouds}
        
\author[1]{Jonatan B. Bastos}
\author[1,2]{Francisco F. Rivera}
\author[2]{Oscar G. Lorenzo}
\author[2]{David L. Vilariño}
\author[1,2]{Jos\'e~C.~Cabaleiro}
\author[1]{Alberto M. Esmorís}
\author[1,2]{Tom\'as F. Pena}

\affil[1]{Centro Singular de Investigaci\'on en Tecnolox\'ias Intelixentes (CiTIUS), Universidade de Santiago de Compostela, Spain}
\affil[2]{Departamento de Electr\'onica e Computaci\'on, Universidade de Santiago de Compostela, Spain}

\begin{abstract}
With the advancement of 3D scanning technologies, point clouds have become fundamental for representing 3D spatial data, with applications that span across various scientific and technological fields. Practical analysis of this data depends crucially on available neighborhood descriptors to accurately characterize the local geometries of the point cloud. This paper introduces LitS, a novel neighborhood descriptor for 2D and 3D point clouds.

LitS are piecewise constant functions on the unit circle that allow points to keep track of their surroundings. Each element in LitS' domain represents a direction with respect to a local reference system. Once constructed, evaluating LitS at \emph{any} given direction gives us information about the number of neighbors in a cone-like region centered around that same direction. Thus, LitS conveys a lot of information about the local neighborhood of a point, which can be leveraged to gain global structural understanding by analyzing how LitS changes between close points. In addition, LitS comes in two versions ('regular' and 'cumulative') and has two parameters, allowing them to adapt to various contexts and types of point clouds. Overall, they are a versatile neighborhood descriptor, capable of capturing the nuances of local point arrangements and resilient to common point cloud data issues such as variable density and noise.
\end{abstract}


\begin{document}

\flushbottom
\maketitle
\thispagestyle{empty}

\section{Introduction}
In recent years, the proliferation of 3D acquisition technologies, such as LiDAR \cite{lidar} and structured light scanners \cite{sls}, has led to a surge in point cloud data's availability and significance. Point clouds are foundational in diverse fields, including \emph{autonomous vehicle navigation} \cite{autonomous_driving}, \emph{robotics} \cite{robotics}, \emph{cultural heritage preservation} \cite{cultural_heritage} and \emph{urban planning} \cite{urban_planning}. And successful analysis of this data hinges on our ability to extract meaningful local geometric attributes that capture the underlying structure of the point cloud \cite{primitives_2020}. The availability of effective local descriptors is therefore essential for successfully carrying out important tasks such as \emph{object recognition} \cite{object_recognition}, \emph{segmentation} \cite{fast_segmentation_2016}, \emph{classification} \cite{classification_2016}, or \emph{registration} \cite{registration_2019}. Especially in the presence of noise, varying density, occlusion and clutter, these are still challenging research problems \cite{still_challenging_1, still_challenging_2}.

The primary motivation behind LitS’ conception was that there are tasks in which it could be handy to know beforehand the polar coordinates of a point’s neighborhood (centered at that point, in a 2D point cloud). Then we could, given some direction of interest from a point $p$, readily compute how many neighbors are close to that direction. However, the polar coordinate of a point with respect to another becomes less relevant the closer the two points are from each other, since getting closer means the radial coordinate approximates zero, which completely determines the position of that point. This motivates giving less weight to closer neighbors, and more to further ones. As a first step, it seems sensible to choose a threshold so that neighbors that are too close to $p$ are ignored, since their polar coordinates will be of relatively low relevance. Moreover, that threshold should be on the same scale as the neighborhood so, rather than specifying it as an absolute value, it makes sense to set it as a proportion $\lambda$ of the neighborhood radius $r$, so that neighbors at a distance lower than $\lambda r$ are ignored. Then we could assign a weight for every remaining neighbor by thinking of $p$ \emph{as a ball} with radius $r_p:=\lambda r$ and of neighbors as \emph{light sources illuminating} $p$'s boundary: each neighbor \emph{lights up} a circular arc, whose relative length prescribes its weight. Indeed, this makes neighbors with a radial coordinate close to $r_p$ receive a low weight, while neighbors further away receive increasingly more, up to a hard upper bound of $\pi r_p$. Going back to our starting premise, we can simply stipulate that a given direction is close to an illuminating neighbor (i.e. a neighbor at least $r_p$ distance away from $p$) if that direction’s representative as an element of $p$’s boundary is lit up. Now, assuming that neighbors illuminate circular arcs uniformly, we can encode a neighbor’s lit-up circular arc as an angular interval of the unit circle $S^1$. And, if $I_k$ denotes the angular interval of the $k$-th illuminating neighbor, then the function $\mathbbm{1}_{\bigcup_kI_k}$ tells us exactly which directions have a neighbor close by and which ones do not. In addition, illuminance levels are additive, so it is natural to consider $\sum_k\mathbbm{1}_{I_k}$ as well, which exactly expresses the number of neighbors that are close to any direction. These two functions are $p$’s regular and cumulative LitS, respectively.

The primary goal of this paper is to serve as an introduction to LitS as a novel point cloud descriptor and introduce some practical applications. Establishing the novelty of a new local descriptor is not feasible, as the presence of some equivalent formulation in prior work cannot be ruled out in practice. Nonetheless, several comprehensive reviews and comparative studies provide a wide overview of existing point cloud descriptors, and none report a method resembling LitS. Foundational benchmark analyses such as Mikolajczyk and Schmid's \cite{mikolajczyk_2005} established unified criteria for evaluating descriptors, while Alexandre \cite{alexandre_2012} compared the main 3D descriptors then implemented in the \emph{Point Cloud Library}—mainly SHOT, PFH, and FPFH—showing that recognition tasks at the time relied largely on these conventional formulations. Eigenvalue-based approaches were first proposed by Weinmann in \cite{weinmann_2015} and later applied by Hackel et al. \cite{fast_segmentation_2016} for segmentation. Although they share LitS’s use of covariance information, their mathematical construction and encoded features are fundamentally different. Normal-based descriptors, systematically compared by Mateo et al. \cite{mateo_etal_2014}, form another separate family that exploits orientation information, but not the angular encoding central to LitS. Broader comparative works such as Guo et al. \cite{guo_etal_2015} confirm that the field of handcrafted local descriptors remains dominated by these established families. Variants of these descriptors have also been used for classification tasks \cite{hughes_2018}, and some have been adapted ad-hoc for specific applications such as \emph{leaf–wood segmentation} \cite{wood_leaf} or \emph{pedestrian detection} \cite{pedestrians}. Several studies also employ polar, spherical, or cylindrical representations \cite{zhou_2019, triess_2019}, which, like LitS, organize local neighborhoods according to angular relationships. In such formulations, geometric information may be encoded analytically or learned automatically, using \emph{convolutional neural networks} for instance. The comprehensive surveys by Han et al. \cite{han_etal_2018, han_etal_2023}—covering both handcrafted and learned descriptors—likewise make no mention of any approach comparable to LitS. While it is not possible to conclusively prove that no similar descriptor has ever been proposed, our review of the literature indicates that LitS constitutes a genuinely novel formulation for describing local geometry in point clouds.


Section~\ref{sec:definition} constructs LitS from a formal mathematical standpoint.
Next, in Section~\ref{sec:properties}, we explore the intrinsic properties of LitS.
Then, Section~\ref{sec:using-lits} starts by leveraging LitS for boundary detection tasks, an example which hopefully can throw some insight into how to apply LitS in general. Lastly, LitS' representation for corner- and line-type neighborhoods is analyzed, relating measurements of the LitS function to the relevant geometric properties of the neighborhood; and some LitS drawings are showcased. 

\section{Definition}\label{sec:definition}
In this section we define LitS formally, taking care to consider parameters in play explicitly and generalizing LitS' definition where possible. We start by stipulating that a neighbor influences or illuminates every direction or region of $p$’s boundary that is visible from its position. Taking into account that later on we assume neighbors illuminate circular arcs uniformly, this is rather arbitrary. So we might want to impose that neighbors light up shorter circular arcs, or maybe even larger ones, which will have an impact on LitS’ smoothness and amount of information about the distribution of the illuminating neighbors’ polar coordinates. This leads to the introduction of the parameter $\varphi$, which allows us to implicitly specify the length of circular arcs lit up by illuminating neighbors. In addition, the last part of this section deals with generalizing LitS to 3D point clouds, which includes a mathematical derivation for constructing them, as well as an alternative definition for the base case $\varphi=\pi/2$.

\begin{wrapfigure}{r}{0.24\textwidth}
\includegraphics[width=0.9\linewidth]{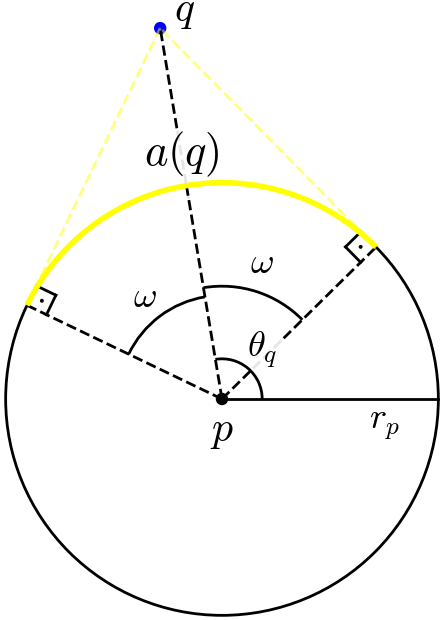}
\caption{Circular arc lit up by $q$.}\label{fig:lit-arc-concept}
\end{wrapfigure}

Let $P$ be a \emph{2D point cloud}, $p\in P$ a fixed point, and $Q$ a neighborhood centered at $p$. For any neighbor $q\in Q$, $q\neq p$, we denote its polar coordinates with respect to $p$ by $r_q$ and $\theta_q$, and define the \emph{radius of} $Q$ as the quantity $r_Q:=\max_{q\in Q}r_q$. Given a value $\lambda\in (0,1]$ we may formally refer to $r_p:=\lambda r_Q$ as the \emph{radius of} $p$, and to $Q_\lambda:=Q\setminus B(p,r_p)$ as $p$'s \emph{illuminating neighbors}. Now, for any neighbor $q\in Q_\lambda$, consider the points of the circumference with center $p$ and radius $r_p$ for which a segment to $q$ does not intersect any other point of the same circumference; i.e., the ones that are "directly visible from $q$". They form a \emph{circular arc}, call it $a(q)$, which can be parameterized as $p+r_p(\cos t, \sin t)\,,\ \theta_q-\omega <t<\theta_q+\omega$, for some $\omega>0$. See Figure~\ref{fig:lit-arc-concept}.

Since both $p$ and $r_p$ are already known and the interesting part of $a(q)$ will be angular information, it makes sense to write $a(q)=(\theta_q-\omega, \theta_q+\omega)$, regarding $a(q)$ as an open interval of the unit circle $S^1$.

Any line meeting a circle at exactly one point must do it tangentially, so the triangle with vertices $p$, $q$ and $p+r_p\big(\cos(\theta_q-\omega), \sin(\theta_q- \omega)\big)$ must be a right triangle. Hence $\cos\omega=r_p/r_q$, and it follows
\begin{equation}\label{eq:a_of_q}
    a(q)=\big(\theta_q-\arccos(r_p/r_q),\theta_q+\arccos(r_p/r_q)\big).
\end{equation}

Notice that, however improbable, it might well happen that $r_q=r_p$, and then (\ref{eq:a_of_q}) will yield the empty interval $a(q)=(\theta_q,\theta_q)=\emptyset$. In that case, let's stipulate that $a(q)=\{\theta\}$. Now, we are ready to define the \emph{LitS of} $p$ as the function \makebox{$L_p[\lambda]:S^1\rightarrow\{0,1\}$}
that assigns $1$ to an angle \mbox{$t\in S^1$} if $t$ belongs to $a(q)$ for any $q\in Q_\lambda$, and $0$ otherwise. Making use of \emph{indicator functions}, this is
\begin{equation}\label{eq:lit-angles}
    L_p[\lambda]:=\mathbbm{1}_A,\ \text{ where }\ A=\bigcup_{q\in Q_\lambda}a(q).
\end{equation}

This completes LitS' definition from a formal standpoint. Moving forward, we might refer to an angle, direction, or circular arc as being \emph{lit up}, meaning that their LitS value is nonzero.

\subsection{Cumulative LitS}\label{subsec:cumulative-lit-angles}
\begin{figure}[h]
\centering
\includegraphics[width=1\linewidth]{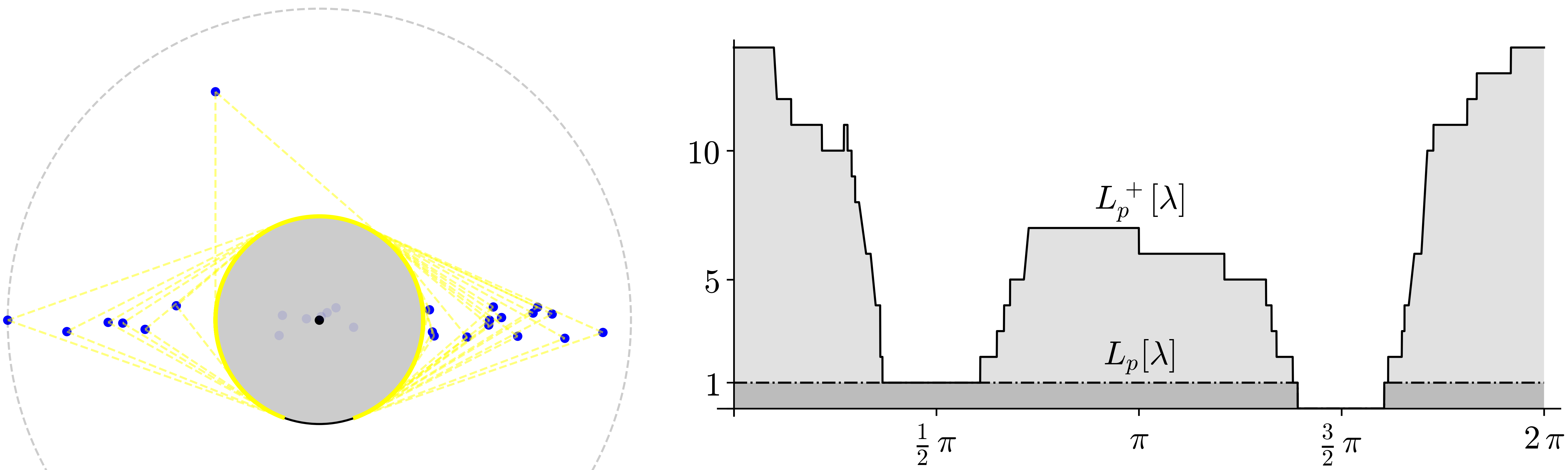}
\caption{Illuminating neighborhood $Q_\lambda$ on the left, with the corresponding plot of LitS and cumulative LitS on the right (dash-dot line and solid line styles, respectively).}
\label{fig:regular-vs-cumulative-lits}
\end{figure}

The more neighbors in $Q_\lambda$, the more likely the circular arcs they light up overlap. The above LitS definition doesn't take note of any overlapping, so it doesn't allow us to distinguish angles lit up by just one neighbor from angles lit up by a large number of neighbors. As a consequence, (regular) LitS do not resist outliers. To circumvent this limitation, we introduce the following \emph{cumulative LitS}:
\begin{equation}\label{eq:cumulative-lit-angles}
    L^+_p[\lambda]:=\sum_{q\in Q_\lambda}\mathbbm{1}_{a(q)}.
\end{equation}
Note that (\ref{eq:cumulative-lit-angles}) is a function from $S^1$ to $\mathbb{N}$. This difference with LitS comes from the fact that LitS tells \emph{whether} an angle is lit up or not, and the cumulative version \emph{how much} an angle is lit up. Now, for any angle $t$ in $S^1$, $L^+_p[\lambda](t)$ counts the number of neighbors that light up $t$, so we can discriminate angles based on their \emph{illuminance}. Clearly, cumulative LitS contain more information about the neighborhood and is much less sensitive to outliers than regular LitS. For instance, Figure~\ref{fig:regular-vs-cumulative-lits} shows a "string neighborhood" with an outlier neighbor alongside regular and cumulative LitS plots. While the former does not accurately describe the neighborhood, a glance at the cumulative LitS plot tells us that we have a horizontal string neighborhood, that there is exactly one outlier in the upper semicircle, and that the right side contains approximately twice as many points as the left side.

\subsection{General Lit Up Arcs}\label{subsec:general-lit-up-arcs}
\begin{wrapfigure}{l}{0.65\textwidth}
\includegraphics[width=0.83\linewidth]{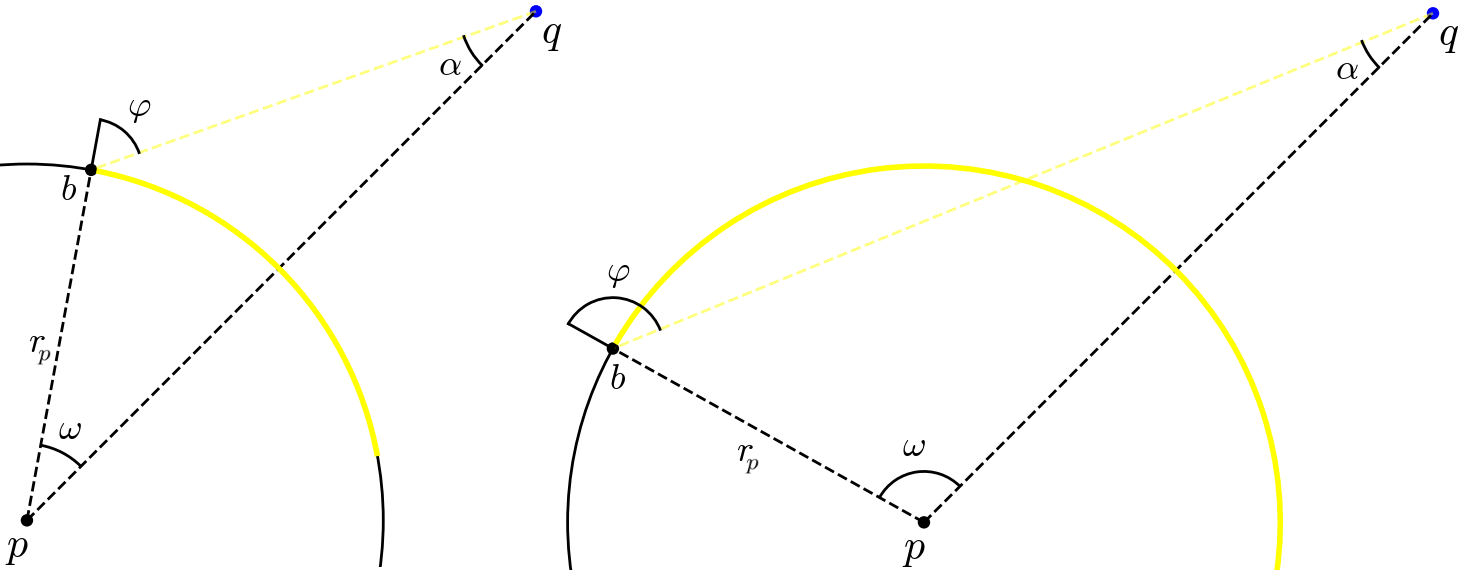}
\caption{Circular arc lit up by $q$ in terms of $\varphi$. Case $\varphi<\pi/2$ on the left, and $\varphi>\pi/2$ on the right. The values used are $\varphi=\pi/3$ and $\varphi=5\pi/7$, respectively. \vspace{-0.8cm}}\label{fig:angle-of-incidence}
\end{wrapfigure}
In (\ref{eq:a_of_q}) we assumed that $q$ lights up every portion of the circular arc that is visible from its position. However, by \emph{Lambert's cosine law}, the illuminance of a given point in that arc will be proportional to the cosine of the \emph{angle of incidence}, which the illuminating ray makes with respect to the point's normal. Thus, it will be maximum at the center, and decrease to $0$ at its ends (which is why $a(q)$ is defined as an open interval). Depending on the application and point density of $P$, we might want to restrict $a(q)$ to angles that are facing $q$ more uprightly, or even expand $a(q)$ beyond angles that are directly visible from~$q$. Since incident rays from $q$ will have a larger angle of incidence the further away from the center of $a(q)$, the natural way to do this is to prescribe a \emph{limiting angle of incidence} $\varphi$ for the ends of $a(q)$. So far we have worked with the case $\varphi=\pi/2$. Figure~\ref{fig:angle-of-incidence} shows both scenarios $\varphi<\pi/2$ and $\varphi>\pi/2$.

\begin{wrapfigure}{r}{0.32\textwidth}
\includegraphics[width=0.9\linewidth]{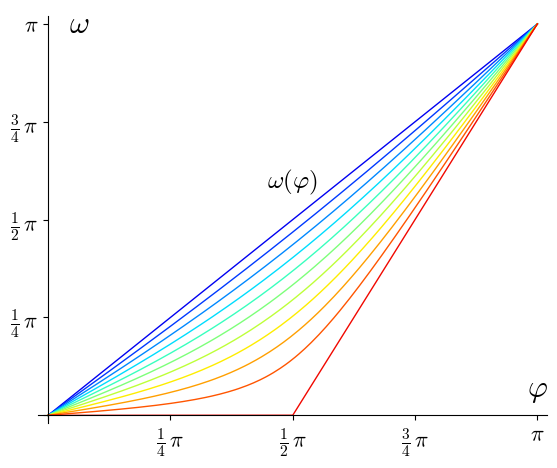}
\caption{$\omega(\varphi)$ with $r_p/r_q$ ranging from $1$ (red) to $0$ (blue) in increments of $0.1$, colored using the \emph{Jet colormap}.}\label{fig:w-of-phi}
\end{wrapfigure}
If $\varphi\neq \pi/2$, then the triangle with vertices $p,q$ and $b$ is no longer a right triangle, so (\ref{eq:a_of_q}) doesn't hold. To come up with a valid expression for this general case apply the \emph{law of sines} to obtain
\begin{equation}\label{eq:law-of-sines}
    \frac{\sin\alpha}{r_p}=\frac{\sin(\pi-\varphi)}{r_q}.
\end{equation}

The interior angles of any triangle add up to $\pi$, so it must be \mbox{$\omega+\alpha+(\pi-\varphi)=\pi$,} from where \mbox{$\alpha=\varphi-\omega$.} This identity, along with \mbox{$\sin(\pi-t)=\sin t$,} $t\in\mathbb{R}$, reads into (\ref{eq:law-of-sines}) as \mbox{$\sin(\varphi-\omega)/r_p=\sin\varphi/r_q$,} and it follows
\begin{equation}\label{eq:w-of-phi}
\omega=\varphi-\arcsin\left(\frac{r_p}{r_q}\sin\varphi\right).
\end{equation}
Figure~\ref{fig:w-of-phi} shows plots of $\omega$ in terms of $\varphi$ for several values of $r_p/r_q$. Equation (\ref{eq:w-of-phi}) generalizes (\ref{eq:a_of_q}) from $\varphi=\pi/2$ to $\varphi\in(0,\pi]$. From here on, we will denote $a_\varphi(q)$ whenever we are using a $\varphi$ other than $\pi/2$. We will also make use of the notation $L_p[\lambda,\varphi]$ and $L^+_p[\lambda,\varphi]$, which implicitly assumes that $\varphi$ stays constant among $p$'s neighbors. Notice, though, that nothing prevents us from specifying different values of $\varphi$ for different neighbors, which could be an effective way to weigh neighbors' contribution to LitS according to a point feature of interest.

\subsection{3D LitS}\label{subsec:3d-lit-angles}
So far we've assumed the underlying point cloud to be two-dimensional. In what follows, assume $P$ is a \emph{3D point cloud}, $p\in P$ is fixed, and $Q$ is a neighborhood centered at $p$. Here we define $r_Q$ as the maximum \emph{radial coordinate} of neighbors in $Q$, which allows us to define $p$'s radius $r_p$ and $p$'s illuminating neighbors $Q_\lambda$ as in the two-dimensional case.
\begin{wrapfigure}[15]{r}{0.36\textwidth}
\includegraphics[width=0.9\linewidth]{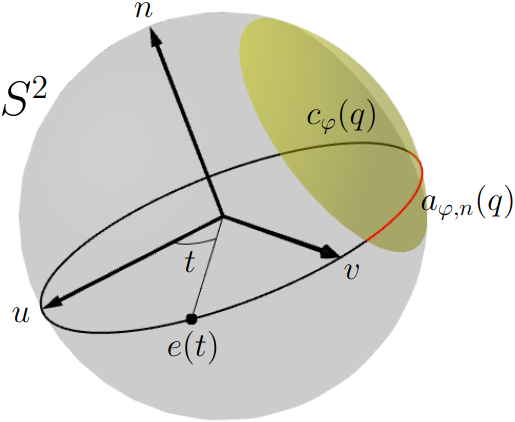}
\caption{LitS' construction along a plane given by $n$.}\label{fig:s2}
\end{wrapfigure}

Our original starting point translates naturally to 3D space: thinking of $q\in Q_\lambda$ as a light source, it makes sense to talk about the region of the \emph{sphere} with center $p$ and radius $r_p$ that is illuminated by incident rays from $q$ with incidence angles lower than the limiting angle of incidence $\varphi$. This will be a (hollow) \emph{spherical cap} $\{p+r_pe:e\in S^2,\ \angle\big(q-(p+r_pe),e\big)<\varphi\},$ where $S^2$ is the \emph{unit sphere} and $\angle(\cdot,\cdot)$ denotes the angle between two vectors in $\mathbb{R}^3$. Notice $q-(p+r_pe)=pq-r_pe$. Ignoring the translation by $p$ and dilation by $r_p$, we consider $c_\varphi(q)$ as a piece of $S^2$;
\begin{equation}\label{eq:c-of-q}
    c_\varphi(q):=\{e\in S^2:\angle(pq-r_pe,e)<\varphi\}.
\end{equation}

Now we can define the \emph{LitS} and \emph{cumulative LitS of $p$} via
\begin{align}
L_p[\lambda,\varphi] &:= \mathbbm{1}_C, \ \ C=\bigcup_{q\in Q_\lambda}c_\varphi(q)\,; \label{eq:lit-angles-3d}\\
L_p^+[\lambda,\varphi] & :=\sum_{q\in Q_\lambda}\mathbbm{1}_{c_\varphi(q)}. \label{eq:cumulative-lit-angles-3d}
\end{align}

Even though (\ref{eq:c-of-q}) is not explicit like (\ref{eq:a_of_q}) we are still able, for any given $e\in S^2$, to compute $L_p[\lambda,\varphi](e)$ and $L_p^+[\lambda,\varphi](e)$ by checking whether $e\in c_\varphi(q)$ for every $q\in Q_\lambda$. Clearly, with (\ref{eq:lit-angles-3d}) we are done as soon as one of those checks holds, but with (\ref{eq:cumulative-lit-angles-3d}) we need to loop over all of $Q_\lambda$. In contrast, (\ref{eq:lit-angles}) and (\ref{eq:cumulative-lit-angles}) allowed us to construct LitS as \emph{piecewise constant} functions for optimal evaluation, which did not need to access $Q_\lambda$ once constructed. We could hope to reformulate (\ref{eq:lit-angles-3d}) and (\ref{eq:cumulative-lit-angles-3d}) to achieve similarly useful definitions. However, a key property that allowed the existence of such a representation in 2D space is that set operations with circular arcs produce circular arcs, which is far from true for spherical caps in 3D space.

Instead of trying to build LitS on a whole sphere, let us take a plane of interest through $p$ and build them along the \emph{great circle} that results from intersecting the sphere with radius $r_p$ centered at $p$ with that plane. Given that we need to perform an eigendecomposition of the neighborhood covariance matrix to ensure LitS are rotationally invariant (see Subsection~\ref{subsec:invariances}), a great default choice would be $p$'s tangent plane, defined as the plane spawned by the two eigenvectors with higher eigenvalues. This approach should be enough for predominantly flat point clouds, where $p$'s tangent or horizontal plane conveys sufficient geometric information about $p$'s surroundings---and where the relative size of the smallest eigenvalue provides a quantitative measure of local planarity. And, for the non-flat ones, nothing prevents us from taking more than one plane, building the corresponding LitS, and combining their information.

Let's assume the plane of choice is defined by a unit normal vector $n$, and let $(u, v)$ be an orthonormal basis for it. Notice that, if we choose $p$'s tangent plane, this basis naturally corresponds to $u = v_1$, $v = v_2$, and $n = v_3$, where $v_1$, $v_2$ and $v_3$ are the eigenvectors of the covariance matrix ordered by decreasing eigenvalue. Our target great circle can then be parameterized by $p+r_p\big(\cos(t)u+\sin(t)v\big)$, $0\leq t<2\pi$. Ignoring the translation and dilation once again we can use $e(t):=\cos(t)u+\sin(t)v$, effectively embedding it into $S^2$. We are interested in the \emph{LitS} and \emph{cumulative LitS of $p$ along $n$}, which we formally define by
\begin{equation}\label{eq:LitS-along-n}
L_{p,n}[\lambda,\varphi](t):=L_p[\lambda,\varphi]\big(e(t)\big) \text{ \ and \ }L_{p,n}^+[\lambda,\varphi](t):=L_p^+[\lambda,\varphi]\big(e(t)\big)\,,\ \,t\in S^1,
\end{equation}
respectively. Now, for every $q\in Q_\lambda$, call $a_{\varphi,n}(q)$ to the arc from $e(S^1)$ that is illuminated by $q$; that is, \mbox{$a_{\varphi,n}(q):=c_\varphi(q)\bigcap e(S^1)$.} See Figure~\ref{fig:s2}. Provided it is nonempty, at its ends the defining condition of (\ref{eq:c-of-q}) will be met with equality, which leads to
\begin{equation}\label{eq:angle-equality-3d}
\cos\varphi=\frac{\big(pq-r_pe(t)\big)\cdot e(t)}{|pq-r_pe(t)|\cdot|e(t)|}=\frac{pq\cdot e(t)-r_p}{|pq-r_pe(t)|}.
\end{equation}
Squaring both sides and using $|w|^2=w\cdot w,\ w\in\mathbb{R}^3,$ the fact that $|pq|=r_q$, and the Pythagorean identity $\cos^2\varphi+\sin^2\varphi=1$, we get
$$
\psi_t^2-2r_p\sin^2(\varphi)\psi_t+r_p^2\sin^2\varphi-r_q^2\cos^2\varphi=0,
$$
where $\psi_t:=pq\cdot e(t)$. This is a second-degree polynomial equation in $\psi_t$, whose solutions are
$$
\psi_t^\pm=r_p\sin^2\varphi\pm|\cos\varphi|\sqrt{r_q^2-r_p^2\sin^2\varphi}.
$$
Although we find two answers for $\psi_t$, the inequality
$$
|\cos\varphi|\sqrt{r_q^2-r_p^2\sin^2\varphi}\geq|\cos\varphi|\sqrt{r_p^2-r_p^2\sin^2\varphi}=r_p\cos^2\varphi
$$
leads to $\psi_t^+\geq r_p$ and $\psi_t^-\leq r_p$. And, since it must be $\text{sgn\,}[\cos\varphi]=\text{sgn\,}[\psi_t-r_p]$ because \mbox{of (\ref{eq:angle-equality-3d}),} we find that $\psi_t=\psi_t^+$ for $\varphi\leq\pi/2$, and $\psi_t=\psi_t^-$ for $\varphi\geq\pi/2$. Taking into account the sign of the $\cos$ function, we can sum this up with
\begin{equation}\label{eq:lambda-t}
\psi_t=r_p\sin^2\varphi+\cos\varphi\sqrt{r_q^2-r_p^2\sin^2\varphi}.
\end{equation}
Expressing $pq$ in terms of the orthonormal basis $(u,v,n)$, $pq=\alpha u+\beta v+\gamma n$, we have
$$
\psi_t=(\alpha u+\beta v+\gamma n)\cdot\big(\cos(t)u+\sin(t)v\big)=\alpha\cos t+\beta\sin t.
$$
Now, any linear combination of cosine and sine with equal periods can be written as a cosine with the same period but different amplitude and a phase shift. Precisely,
\begin{equation}\label{eq:3d-solutions}
\psi_t=A_q\cos(t-\phi_q),\ \ A_q=\sqrt{\alpha^2+\beta^2},\ \ \phi_q=\arctan(\beta/\alpha).
\end{equation}
This equation can have 
\begin{itemize}
\item[$\bullet$] infinitely many solutions, if $\psi_t=A_q=0$. This means that $q$ lies on the line that passes through $p$ with direction $n$, and $r_q=|\gamma|=r_p\tan(\pi-\varphi)$ so that $c_\varphi(q)$'s boundary coincides exactly with $e(S^1)$. Even though every $t\in S^1$ is a solution, $c_\varphi(q)$ is open, so no arc from $e(S^1)$ gets lit up; i.e., $a_{\varphi,n}(q)=\emptyset$.
\item[$\bullet$] no solutions, if $|\psi_t|>A_q$. Then $c_\varphi(q)$'s boundary does not intersect $e(S^1)$, which implies that $c_\varphi(q)$ contains none of $e(S^1)$, or all of it. Since $\Phi:=\pi-\arctan\big(r_q/r_p\big)$ is the value at which $c_\Phi(q)$ becomes a semisphere of $S^2$, it follows $a_{\varphi,n}(q)=\emptyset$ if \mbox{$\varphi<\Phi$,} and $a_{\varphi,n}(q)=e(S^1)$ if $\varphi>\Phi$.
\item[$\bullet$] one solution $t_0$, if  $\psi_t=A_q\neq0$. Here $c_\varphi(q)$'s boundary intersects $e(S^1)$ at exactly one point; $e(t_0)$. We have $a_{\varphi,n}(q)=\emptyset$ if $\varphi<\Phi$, and $a_{\varphi,n}(q)=e(S^1\setminus\{t_0\})$ if $\varphi>\Phi$.
\item[$\bullet$] two solutions $t_0$ and $t_1$, otherwise. Call $I(\eta_0,\eta_1)$ to the open angle-interval that starts at $\eta_0$ and ends at $\eta_1$, swept counterclockwise. Then we will have $a_{\varphi,n}(q)=e\big(I(t_0,t_1)\big)$ if $\arctan2(\beta,\alpha)$ lies in $I(t_0,t_1)$, and $a_{\varphi,n}=e\big(I(t_1,t_0)\big)$ otherwise.
\end{itemize}

Notice $A_q$ and $\arctantwo(\beta,\alpha)$ are the polar coordinates of $pq$'s orthogonal projection onto our plane of choice, measured counterclockwise from $u$. Much like in the 2D scenario, this analysis allows us to construct $L_{p,n}[\lambda,\varphi]$ and $L_{p,n}^+[\lambda,\varphi]$ as piecewise constant functions through $e$'s parameter space.

Finally, it should be noted that there's a simpler alternative definition of 3D LitS that could be equally useful, particularly for mostly flat point clouds and $\varphi\approx\pi/2$. This alternative approach is motivated by the idea that we could ignore one spatial coordinate in $P$ and calculate LitS as if it were a 2D point cloud. While this would only allow us to compute LitS along the three coordinate planes, it already hints at how the previous analysis on the $XY$-plane could be carried over to any plane through $p$: project every point in $Q_\lambda$ orthogonally onto the plane, choose an orthonormal basis on it so that we can calculate distances and measure angles, and mimic LitS 2D-space construction. For $\varphi=\pi/2$, it can be shown that this alternative notion of LitS coincides with the one we just introduced in (\ref{eq:LitS-along-n}), and it is more computationally efficient. However, except for $\varphi=\pi/2$, it is not consistent with (\ref{eq:lit-angles-3d}) and (\ref{eq:cumulative-lit-angles-3d}): If $\varphi>\pi/2$, then there could be a point $q$ with $A_q<r_p$ and still have $a_{\varphi,n}(q)\neq\emptyset$. So $q$ would necessarily be ignored towards LitS' computation (since its projection falls at a distance smaller than $r_p$ from $p$), despite lighting up an arc from $e(S^1)$. Moreover, if $q$ is a point for which $A_q\geq r_p$ and $\gamma\neq0$, then $a_{\phi,n}(q)=\emptyset$ for $\varphi$ small enough, but it would still contribute to LitS' computation.

\section{Properties}\label{sec:properties}
Here we start by analyzing the meaning of a LitS value. That is, what can we infer about $p$’s neighborhood just by evaluating LitS in some direction? As it turns out, this question has a nice geometrical interpretation. Then, we analyze edge cases for LitS’ parameters $\lambda$ and $\varphi$, and we show LitS are invariant to translations, rotations and dilations, ensuring consistency under geometry-preserving space transformations. Next, we introduce a measure of surroundedness for points, called $\varphi$-surroundedness. Lastly, we show that cumulative LitS on 3D point clouds are a neighborhood transform, in the sense that it is possible to retrieve $p$’s neighborhood from them.

While Section~\ref{sec:definition} outlines the definition and construction of LitS, this one focuses on their properties, which already provide a partial answer to why they could be a useful asset in the analysis of both 2D and 3D point clouds. We defined $L_p[\lambda,\varphi]$ and $L_p^+[\lambda,\varphi]$ as the LitS and cumulative LitS at a point $p$ from a point cloud $P$, respectively. The radius $r_p$ specifies the distance range for $p$'s illuminating neighborhood $Q_\lambda$, and $\varphi$ the limiting angle of incidence from an illuminating neighbor to a point in the circle or sphere centered at $p$ with radius $r_p$ to be lit up. Notice LitS' domain differs from 2D to 3D, but our notation is dimension-agnostic, which allows us to refer to LitS as a unified object. Due to our inability to build 3D LitS on the whole $S^2$, we devoted most of Subsection~\ref{subsec:3d-lit-angles} to construct LitS along a normal vector $n$, which has domain $S^1$ and a different notation. However, as made explicit by \eqref{eq:LitS-along-n}, they are LitS' \emph{restriction} to the great circle resulting from intersecting $S^2$ with the plane determined by $n$, which is indeed part of the same object.

\subsection{Meaning of a LitS value}\label{subsec:visible-regions}
\begin{wrapfigure}{l}{0.55\textwidth}
\includegraphics[width=0.9\linewidth]{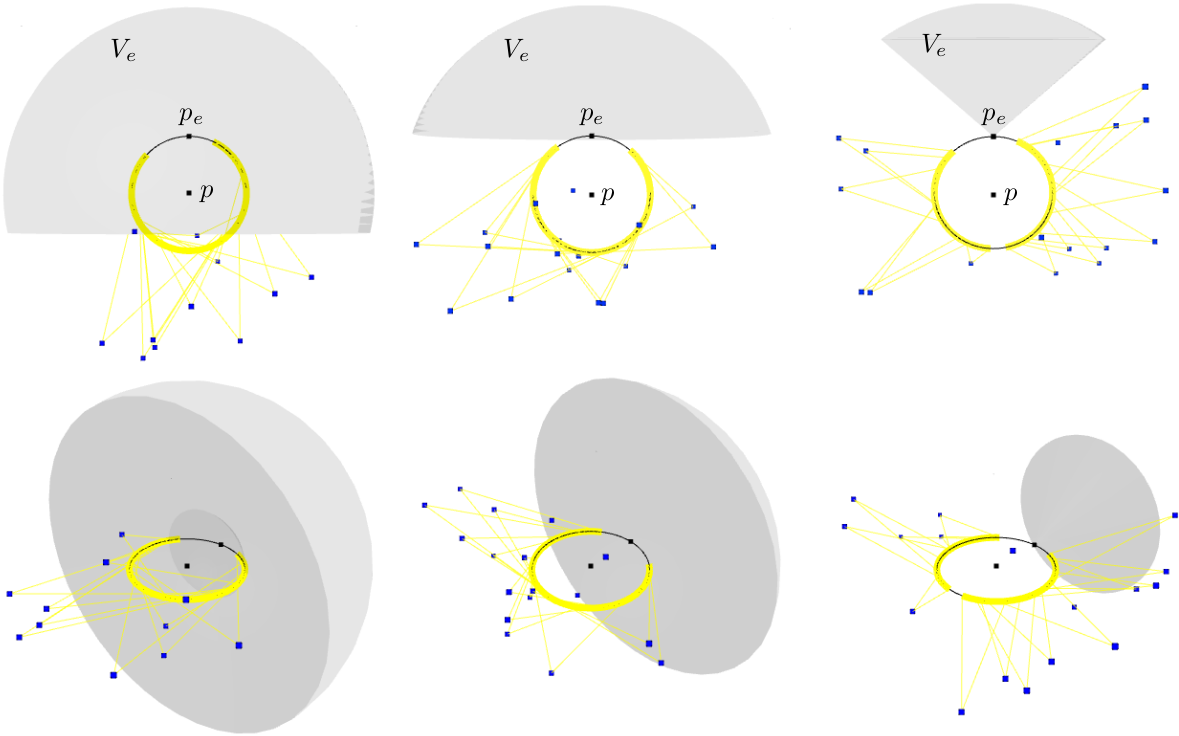}
\caption{Top view (top) and high-angle 3/4 view (bottom) of three $\varphi$-visible regions from an unlit element $e\in S^2$ with varying values of $\varphi$: $2\pi/3$ (left), $\pi/2$ (center) and $\pi/4$ (right). Points in $Q_\lambda$ are shown in blue, and LitS are calculated along a normal vector.\vspace{-0.4cm}}\label{fig:visible-regions}
\end{wrapfigure}
Assume $P$ is a 3D point cloud and, for a given $e$ in $S^2$, we have $L_p^+[\lambda,\varphi](e)=m$. Then, by definition, there will be exactly $m$ neighbors in $Q_\lambda$ from which $p_e:=p+r_pe$ is visible with incident angles less than $\varphi$. This means that the $\varphi$\emph{-visible region from $e$},
\begin{equation*}
V_e:=\left\{x\in B(p,r_Q)\setminus B(p,r_p):\angle(x-p_e,e)<\varphi\right\},
\end{equation*}
where $B(y,z)$ denotes the open ball centered at $y$ with radius $z$ in $\mathbb{R}^3$, contains exactly $m$ points from $P$. Thus
\begin{equation*}\label{geometric-lit-angles}
L_p^+[\lambda,\varphi](e)=\big|V_e\cap P\big|\,,\ e\in S^2.
\end{equation*}
And essentially the same argument shows that, for every $e\in S^2$, $L_p[\lambda,\varphi](e)$ evaluates to 0 if \mbox{$V_e\cap P=\emptyset$,} and to 1 otherwise. While they do not provide another way of constructing LitS, these alternative representations allow us to extract useful geometric information about the surroundings of~$p$. Figure~\ref{fig:visible-regions} depicts three scenarios of an element $e$ whose LitS value is $0$ along with its corresponding region $V_e$, necessarily containing no point \mbox{from $P$.}\\

\begin{wrapfigure}[21]{r}{0.27\textwidth}
\includegraphics[width=1.0\linewidth]{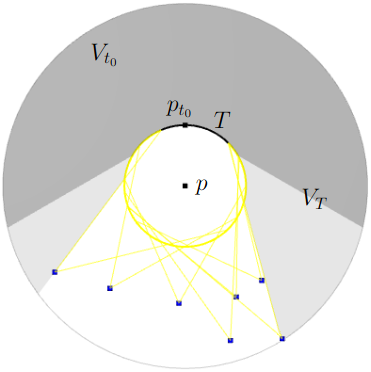}
\caption{Comparison between $V_{t_0}$, the $\varphi$-visible region from an angle $t_0$, in dark grey; and $V_T$, the union of all $\varphi$-visible regions from angles in~$T$, in light and dark grey. The parameters used are $\lambda=1/3$ and $\varphi=2\pi/3.$}\label{fig:maximal-visible-region}
\end{wrapfigure}

The 2D case is entirely analogous, and switching '$e$' for '$t$' and '$S^2$' for '$S^1$' above yields a correct analysis. In addition, the top row of Figure~\ref{fig:visible-regions} with Figure~\ref{fig:maximal-visible-region} in place of the left drawing provides the corresponding figure of interest for $\varphi$-visible regions. 

The fact that we can construct LitS as constant piecewise functions means that the cost of evaluating a single LitS value is essentially the same as the cost of evaluating the maximal constant interval around it. A remarkable consequence is that we get broader visible regions at no extra cost. For example, say we have constructed LitS for a point $p$ in a 2D point cloud $P$. Then, upon evaluating at some $t_0$ and getting $L_p[\lambda,\varphi](t_0)=0$, we know $V_t$ contains no point from $P$. However, because $L_p[\lambda,\varphi]$ is piecewise constant, the same computation could give us a maximal interval $T\subset S^1$ containing $t_0$ for which $L_p[\lambda,\varphi](t)=0,\ t\in T$, and it follows that $V_T:=\bigcup_{t\in T}V_t$ contains no point from $P$. As shown in Figure~\ref{fig:maximal-visible-region}, $V_T$ can be substantially larger than $V_t$.

\subsection{Limiting Behavior for $\lambda$ and $\varphi$}\label{subsec:limiting-behavior}

The parameter $\lambda$ has the intuitive meaning of being the percentage of $Q$'s radius that splits the neighborhood into ignored neighbors, whose radial coordinate is below $\lambda\%$ of $r_Q$; and illuminating ones, which actually contribute to LitS construction. Notice that, in the definition of LitS at the beginning of Section~\ref{sec:definition}, $\lambda$ is allowed to assume any value in $(0,1]$. When $\lambda=1$ it is easy to see that $Q_\lambda$ consists of neighbors (most likely a single one) which attain the maximum radial coordinate $r_Q$, and that then LitS will simply be the indicator function of the angular (directional in the 3D scenario) coordinates of those neighbors.

The situation at $0$ is a bit more convolved since $\lambda=0$ would imply $r_p=\lambda r_Q=0$, and then the parameterization of circular arcs via $p+r_p(\cos t,\sin t)$ collapses into a single point. For 2D point clouds, recall that $a_\varphi(q)=\left(\theta_q-\omega,\theta_q+\omega\right)$, where $\omega$ is given by \eqref{eq:w-of-phi}. Notice that $\omega$ depends on $r_p$, which in turn depends on $\lambda$. From the continuity of the $\arcsin$ function it follows that $\omega\to\varphi$ as $\lambda\searrow0$, allowing us to define $a_\varphi(q)=\left(\theta_q-\varphi,\theta_q+\varphi\right)$ for $\lambda=0$. This means that in the resulting LitS, given by (\ref{eq:lit-angles}) and (\ref{eq:cumulative-lit-angles}), any neighbor $q$ contributes with a patch of constant length $2\varphi$ centered at $\theta_q$, regardless of its distance $r_q$ to $p$. For 3D point clouds, observe that $pq-r_pe$ converges to $pq$ as $\lambda\searrow0$. The continuity of the $\arccos$ function, the dot product and the norm in $\mathbb{R}^3$ ensure that we can take limits in the defining condition of \eqref{eq:c-of-q} to get $c_\varphi(q)=\{e\in S^2:\angle(pq,e)<\varphi\}$ for $\lambda=0$. This is a spherical cap centered at $q$'s direction from $p$ and constant angular aperture $2\varphi$, in agreement with the 2D case.

Section~\ref{subsec:general-lit-up-arcs} introduced the \emph{limiting angle of indicence}; a parameter $\varphi$ that generalizes the definition of the circular arc lit up by an illuminating neighbor. As Figure~\ref{fig:angle-of-incidence} suggests, the construction of that subsection makes sense for $0<\varphi\leq\pi$. However, it might be handy to complete this range to $[0,+\infty]$. To do so, it will be enough to define $a_0(q)=\{\theta_q\}$ and $a_\varphi(q)=S^1$ for $\varphi>\pi$ in the 2D case; and $c_0(q)=\{pq/|pq|\}$ and $c_\varphi(q)=S^2$ for $\varphi>\pi$ in the 3D case.

\subsection{Translation, Rotation and Scaling Invariances}\label{subsec:invariances}
Let $P$ be a point cloud and, for any given $p\in P$, denote the local neighborhood around $p$ by $Q$. We say a neighborhood descriptor $D=D(p,Q)$ is \emph{invariant} under a family of transformations $\{f_\gamma\}_{\gamma\in\Gamma}$ if
$$
D(f_\gamma(p), f_\gamma(Q)) = D(p,Q),\ \,\gamma\in\Gamma.
$$
Here the $f_\gamma$'s are functions from $P$'s underlying Euclidean space ($\mathbb{R}^2$ or $\mathbb{R}^3$) to itself, and $f_\gamma(Q)$ is the \emph{image set} of $Q$ under $f_\gamma$. In particular, $D$ is called \emph{translation, rotation} or \emph{scaling invariant} if $D$ is invariant under $\{x\mapsto x+t\}_{t\in\mathbb{R}^{2\backslash3}}$\,, $\{x\mapsto Rx\}_{R\in SO(2\backslash3)}$ or $\{x\mapsto kx\}_{k\in\mathbb{R}}$, respectively. Translation invariance allows a descriptor to recognize features irrespective of their position, rotation invariance ensures recognition from any angle, and scaling invariance guarantees feature recognizability is not hindered by scale. These invariances are needed for consistent characterization of features within a scene regardless of their location, orientation or size. Ultimately, they enhance the reliability, generalizability, and efficiency of pattern recognition systems.

LitS are constructed from illuminating neighbors' relative position to $p$. Since relative position is invariant to translation, LitS naturally inherit this trait.

Given a rotation $R_\beta$ in $\mathbb{R}^2$, where $\beta\in S^1$ is the angle of rotation, and any neighbor $q\in Q_\lambda$, we have
\begin{equation*}\label{eq:rotated-arcs}
a(R_\beta q)=(\theta_q+\beta-\omega, \theta_q+\beta+\omega)=\beta+(\theta_q-\omega,\theta_q+\omega)=\tau_\beta(a(q)),
\end{equation*}
where $\tau_\beta$ denotes a shift by $\beta$ radians in $S^1$. Having into account that translations are distributive over unions, it follows from here that
\begin{equation}\label{eq:rotated-arcs-union}
A:=\bigcup_{q\in R_\beta Q_\lambda}a(q)=\bigcup_{q\in Q_\lambda}a(R_\beta q)=\bigcup_{q\in Q_\lambda}\tau_\beta\big(a(q)\big)=\tau_\beta\left(\bigcup_{q\in Q_\lambda}a(q)\right)
\end{equation}
Now, rotations preserve distances, and so $(R_\beta Q)_\lambda=R_\beta Q_\lambda$, which leads to 
$$
L_{p,R_\beta Q}[\lambda,\varphi]=\mathbbm{1}_A=\mathbbm{1}_{\tau_\beta\left(\bigcup_{q\in Q_\lambda}a(q)\right)}=\left(\mathbbm{1}_{\bigcup_{q\in Q_\lambda}a(q)}\right)\circ\tau_\beta=L_{p,Q}[\lambda,\varphi]\circ\tau_\beta.
$$
This shows that a rotation of the neighborhood results in a shift in LitS domain by the same angle, so LitS are rotationally invariant, but only up to a constant shift. And essentially the same argument, in conjunction with the right-distributivity of composition over addition, proves that $L^+_{p,R_\beta Q}[\lambda,\varphi]=L^+_{p,Q}[\lambda,\varphi]\circ\tau_\beta$, so the same conclusion applies for cumulative LitS.

While this weak version of rotational invariance should suffice for applications relying on qualitative measures of LitS' shape or applications that consider a single view of a point cloud, it will not be enough for tasks that require matching neighborhoods such as point cloud registration. To get around this limitation we can use the \emph{covariance matrix} of the neighborhood to define a reference angle intrinsically: Let $Q$ be a neighborhood for which we wish to compute LitS, and denote the matrix whose rows are the Cartesian coordinates of the points in $Q$ by $X$. Then $C_Q:=X^TX$ is a multiple of the covariance matrix of $Q$ centered at~$p$. Covariance matrices are always symmetric and positive semidefinite, which ensures diagonalizability with non-negative eigenvalues. The only case where the two eigenvalues are zero corresponds to the degenerate case where $Q$ consists only of copies of $p$, so we can assume $C_Q$ to have a maximal non-zero eigenvalue. Now, any non-zero scalar multiple of an eigenvector is still an eigenvector for the same eigenvalue, so every maximal nonzero eigenvalue gives two potential reference angles. From here to univocally determine a reference angle it suffices to choose one that minimizes cumulative angular distances to points in~$Q$. While such a minimizer may not be unique, cases of non-uniqueness are extremely rare in practice and typically involve configurations close to symmetric.

To see why the above choice of reference angle is invariant under rotations consider the rotated neighborhood $Q'=R_\beta Q$. The columns of $R_\beta X^T$ contain the coordinates of its neighbors, so the corresponding covariance matrix will be
$$
C_Q':=\big(R_\beta X^T\big)\big(R_\beta X^T\big)^T=R_\beta X^T\big(X^T\big)^TR_\beta^T=R_\beta X^TXR_\beta^{-1},
$$
where the last equality uses the fact that transposition is an idempotent operator, and that the transpose of an orthogonal transformation matches its inverse. Now, if $v$ is an eigenvector of $C_Q$ with eigenvalue $\kappa$, then
$$
C_Q'\big(R_\beta v)=R_\beta X^TXR_\beta^{-1}\big(R_\beta v\big)=R_\beta X^TXv=R_\beta\kappa v=\kappa\big(R_\beta v\big);
$$
i.e., $R_\beta v$ is an eigenvector of $C_Q'$ with the same eigenvalue. As a consequence, our potential reference angles rotate with the neighborhood. Since minimizers of cumulative angular distances also rotate with the neighborhood, this shows that our choice of reference angle is rotationally invariant. In conclusion, if (strong) rotation invariance is required for LitS, then compute the reference angle prior to LitS calculation and express neighbors' angular coordinates with respect to it, which in practice means subtracting the reference angle from each neighbor's angular coordinate. Observe that the above analysis is done for a 2D ambient space, but that the same rationale can be applied to the 3D case with obvious modifications.

Lastly, let's tackle scale invariance in two dimensions: Dilations preserve angles, so the angular coordinates $\theta_q$ remain constant, but a dilation changes the quantities $r_q$, $r_Q$, $r_p$, and $\omega$. Let's decorate those with a prime symbol to tell them apart. Given a constant $k>0$, the rescaled neighborhood $kQ$ has radius $r_{kQ}=kr_Q$, which means that the new radius of $p$ is
$$
r_p'=\lambda r_{kQ}=\lambda kr_Q=kr_p.
$$
In turn, this leads to
$$
\omega'=\varphi-\arcsin\left(\frac{r_p'}{r_q'}\sin\varphi\right)=\varphi-\arcsin\left(\frac{kr_p}{kr_q}\sin\varphi\right)=\omega,
$$
which shows that lit up arcs remain constant under changes of scale, and LitS invariance under dilations follows. Again, the 3D argument is completely analogous.

\subsection{A Surroundedness Measure for Points}\label{subsec:surroundedness}
We can intuitively tell how surrounded $p$ is by points in $Q_\lambda$. Building on the idea that $p$ will be more surrounded the lower the value of $\varphi$ that's needed so that $Q_\lambda$ completely lights up $p$, LitS provide a way of measuring this intuitive surroundedness notion.

We say that a point $p$ with a fixed set $Q_\lambda$ of illuminating neighbors is $\varphi$-surrounded if $L_p[\lambda,\varphi]=1$. Despite this definition being valid regardless of the dimension, the fact that we are unable to deal with 3D LitS as a whole would force us to restrict ourselves to 3D LitS along a normal vector. In practice, the most remarkable difference between these and 2D LitS is that arcs are calculated via \eqref{eq:lambda-t} and \eqref{eq:3d-solutions} instead of \eqref{eq:w-of-phi}. Therefore, let us assume $P$ is a 2D point cloud throughout this subsection for clarity.

\begin{figure}[h]
\centering
\includegraphics[width=1\linewidth]{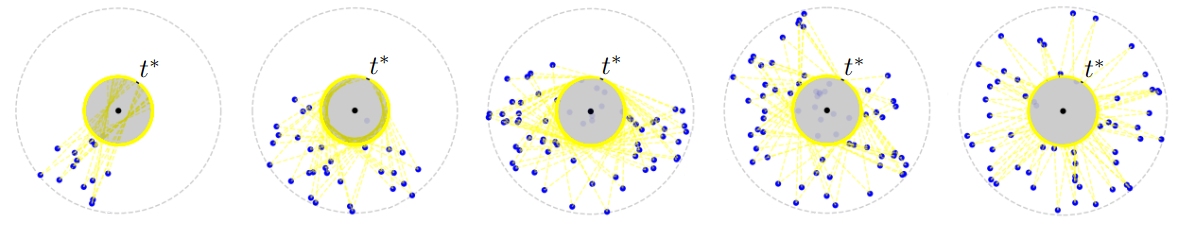}
\caption{Examples of a point and its illuminating neighbors yielding different values of $\varphi^\ast$; from left to right: 90\%, 70\%, 50\%, 30\% and 10\% of $\pi$.}
\label{fig:phi-of-p}
\end{figure}

No value of $\varphi$ will make $p$ $\varphi$-surrounded if $Q_\lambda=\emptyset$, so assume $Q_\lambda\neq\emptyset$. Observe that $a_\varphi(q)$ gets wider as $\varphi$ increases, for every $q\in Q_\lambda$, and so $A_\varphi=\bigcup_{q\in Q_\lambda}a_\varphi(q)$ does too. This implies that if $p$ is $\varphi_0$-surrounded for some $\varphi_0$, then $p$ must be $\varphi$-surrounded for every $\varphi>\varphi_0$. Thus
\begin{equation*}\label{eq:surroundedness-measure}
\varphi^\ast:=\inf\{\varphi:p\text{ is $\varphi$-surrounded}\}
\end{equation*}
is the value we are interested in since it tells us that $p$ is not $\varphi$-surrounded for $\varphi<\varphi^\ast$, and $p$ is $\varphi$-surrounded for $\varphi>\varphi^\ast$. Notice that the infimum is well-defined because $p$ will be $\varphi$-surrounded for $\varphi>\pi$.  In addition, the fact that the $a_\varphi(q)$'s are open along with the continuity of the mappings $\varphi\mapsto|a_\varphi(q)|$ can be used to conclude that if $p$ is $\varphi$-surrounded, then $p$ must be $(\varphi-\varepsilon)$-surrounded for some $\varepsilon>0$. Hence $p$ is not $\varphi^\ast$-surrounded, and we may characterize $\varphi^\ast$ as the highest $\varphi$ for which $p$ is not $\varphi$-surrounded. In particular, this means that the set $S^1\setminus A_{\varphi^\ast}$, most likely a singleton $\{t^\ast\}$, can be interpreted as $p$'s \emph{outside directions} with respect to $Q_\lambda$. See Figure~\ref{fig:phi-of-p}.

To see how $\varphi^\ast$ provides a way to characterize how surrounded $p$ is, note that a low value of $\varphi^\ast$ means $S^1$ can be covered by small circular arcs, whose centers are the directions of points in $Q_\lambda$. Thus, for every possible direction $t\in S^1$, there must be a point from $Q_\lambda$ whose direction from $p$ is close to $t$; i.e., $p$ is heavily surrounded by $Q_\lambda$. And conversely, a high value of $\varphi^\ast$ means big circular arcs don't cover $S^1$, so there must be some uncovered $t\in S^1$, that's necessarily far direction-wise from every point in $Q_\lambda$.

\begin{wrapfigure}{l}{0.5\textwidth}
\includegraphics[width=0.75\linewidth]{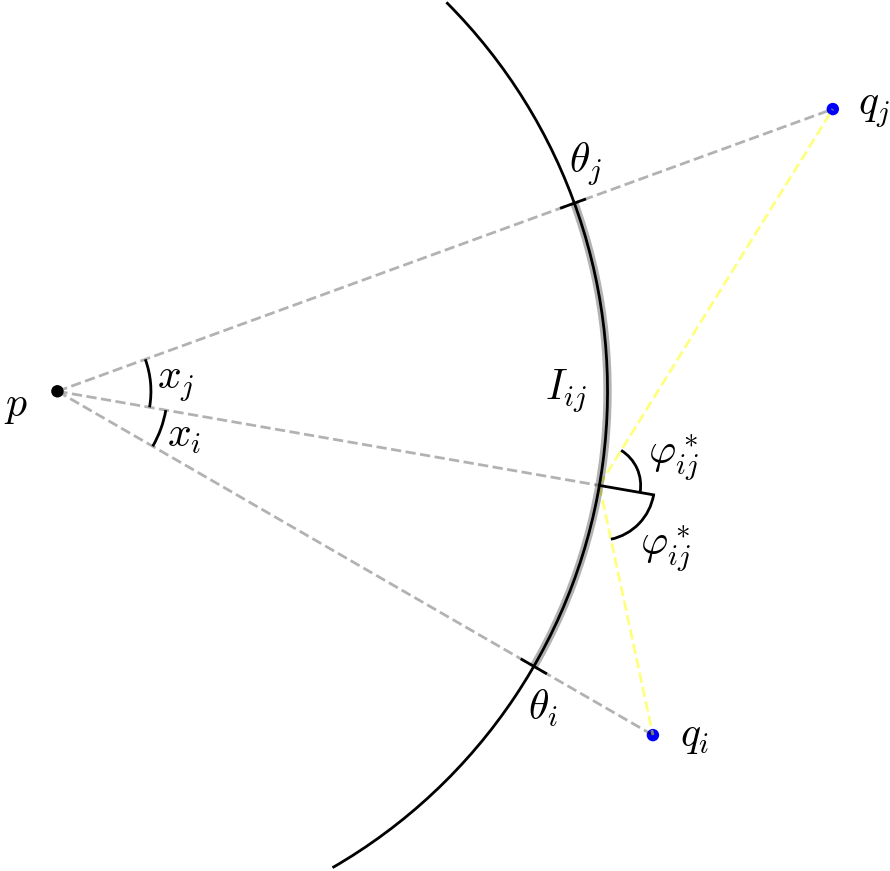}
\caption{Graphical representation of $I_{ij}$, $\varphi^\ast_{ij}$, $x_i$ and $x_j$.\vspace{-0.8cm}}\label{fig:phiij}
\end{wrapfigure}
\leavevmode\\
While we have unequivocally defined $\varphi^\ast$ for a given point and respective set of illuminating neighbors, we still have no means of actually computing it other than successive approximation via trial and error. To this end, a brute-force type algorithm is outlined below.

Assume all points $q_0,q_2,\ldots,q_{n-1}$ in $Q_\lambda$ are arranged attending to their polar coordinates, centered at $p$, counterclockwise and starting at $0$. As further points illuminate wider arcs, if some of the points happen to have the same angular coordinate, then only one of them that is furthest away from $p$ will be relevant, and we may discard the others. Fix a pair $(q_i,q_j)$, with $i\neq j$, and let $I_{ij}$ be the interval of $S^1$ that starts at $\theta_i$ and ends at $\theta_j$, swept counterclockwise. Let $\varphi^\ast_{ij}$ be the value of $\varphi$ for which $a_\varphi(q_i)\cup a_\varphi(q_j)$ covers all of $I_{ij}$ except for one point. According to \eqref{eq:w-of-phi}, we have
\begin{equation*}
    x_k=\varphi^\ast_{ij}-\arcsin\left(\frac{r_p}{r_k}\sin\varphi^\ast_{ij}\right),\ k=i,j\,,
\end{equation*}
for some $x_i$ and $x_j$. See Figure~\ref{fig:phiij}. Because of how $\varphi^\ast_{ij}$ is defined, it must be $x_i+x_j=|I_{ij}|$ so, adding both equations and rearranging terms,
\begin{equation}\label{eq:phi^*_ij}
2\varphi^\ast_{ij}-|I_{ij}|=\arcsin\left(\frac{r_p}{r_i}\sin\varphi^\ast_{ij}\right)+\arcsin\left(\frac{r_p}{r_j}\sin\varphi^\ast_{ij}\right).
\end{equation}
Clearly $|I_{ij}|=\theta_j-\theta_i$ if $i<j$, and $|I_{ij}|=2\pi-\theta_i+\theta_j$ if $i>j$, so (\ref{eq:phi^*_ij}) has $\varphi^\ast_{ij}$ as its only unknown. One can go about taking sines on both sides, using trigonometric addition formulas, the identity $\cos(\arcsin t)=\sqrt{1-t^2}$, and squaring to get rid of radicals, but it soon becomes clear that (\ref{eq:phi^*_ij}) is best solved numerically. To this end, notice that we can easily compute the angles of incidence of rays from $q_i$ and $q_j$ to each other’s projections onto the circumference centered at $p$ with radius $r_p$. The minimum of those angles, call it $\Phi_{ij}$, will be an upper bound for the solution $\varphi_{ij}^\ast$, which will be unique in $[0,\Phi_{ij}]$. Considering also that (\ref{eq:phi^*_ij}) is differentiable and that $\Phi_{ij}$ is actually a good initial guess, the \emph{Newton-Raphson method} would be a strong choice due to its fast (cuadratic) convergence.

Having calculated $\varphi^\ast_{ij}$ for every pair $(q_i,q_j)$, $i\neq j$, consider an interval $I_k:=I_{k,k+1}$, where $k\in\{0,\ldots,n-1\}$ and it is understood that $k+1=0$ if $k=n-1$. By construction, the quantity 
\begin{equation}\label{eq:phi^*_k}
\varphi^\ast_k:=\min\left\{\varphi^\ast_{ij}:i,j\in\{0,\,\ldots\,,n-1\},i\neq j, I_k\subset I_{ij}\right\}
\end{equation}
is the lowest value of $\varphi$ that's required so that $I_k$ is completely illuminated, except for a single point. Therefore it must be that
\begin{equation*}\label{eq:phi^*}
\varphi^\ast=\max_{0\leq k<n}\varphi^\ast_k.
\end{equation*}
As a computational side note, observe that
\begin{equation*}\label{eq:inclusion-equivallence}
I_k\subset I_{ij}\Longleftrightarrow \left\{\begin{array}{lll}
i\leq k<j,            & \text{if} & i<j \\
i\leq k\text{ or }k<j & \text{if} & i>j,\end{array}\right.
\end{equation*}
and that exactly one of $I_k\subset I_{ij}$ and $I_k\subset I_{ji}$ must hold. Thus, calculating the $\varphi^\ast_{ij}$'s in pairs $(\varphi^\ast_{ij},\varphi^\ast_{ji}),\ i<j$, we can find each $\varphi^\ast_k$ by looping over them, updating the minimum in (\ref{eq:phi^*_k}) with the first entry whenever $i\leq k<j$, and with the second otherwise.\\

So far in this subsection, we have used regular LitS to introduce the concept of $\varphi$-surroundedness, which we reduced to the quantity $\varphi^\ast$. For the cumulative counterpart, we say, for a given point $p$ and set of illuminating neighbors $Q_\lambda$, that $p$ is $\varphi$-surrounded \emph{with multiplicity $i$} if $L^+_p[\lambda,\varphi]\geq i$, where $i\geq1$ is an integer. This definition can also be reduced to knowing the highest $\varphi$ for which $p$ is not $\varphi$-surrounded with multiplicity $i$, but we have no direct algorithm for its computation.

\subsection{Cumulative LitS as a Neighborhood Transform}\label{subsec:transformer}
\begin{figure}
\centering
\begin{minipage}{.4\textwidth}
  \centering 
  \includegraphics[width=.9\linewidth]{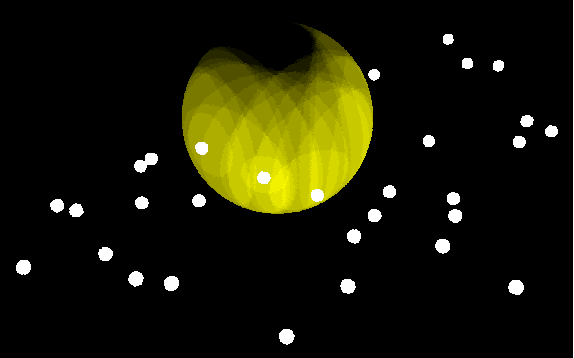}
  \captionof{figure}{3D cumulative LitS of a point with illuminating neighbors in white and encoded values proportional to yellow intensity.}
  \label{fig:cumulative-lits-3d}
\end{minipage}
\begin{minipage}{0.05\textwidth}
$\ $
\end{minipage}
\begin{minipage}{.4\textwidth}
  \centering
  \includegraphics[width=.9\linewidth]{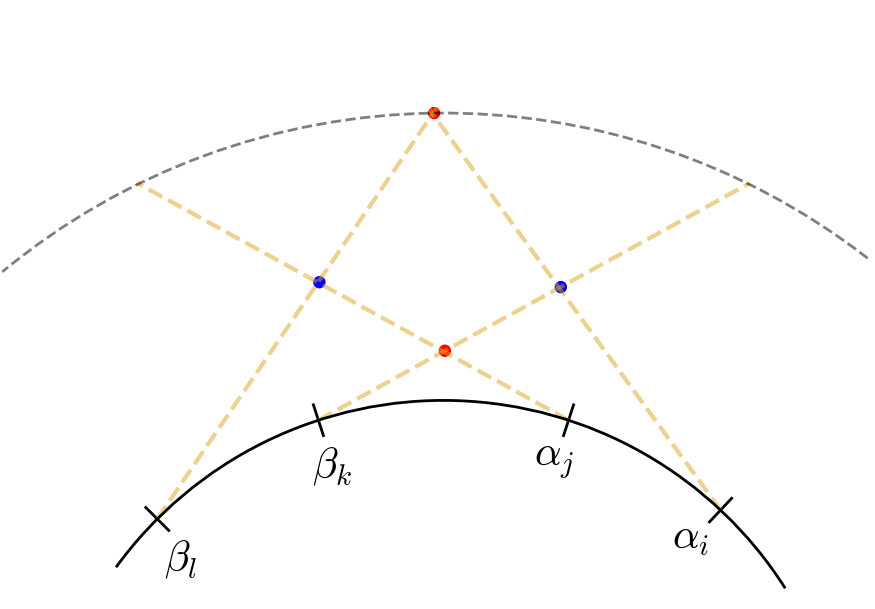}
  \captionof{figure}{Two distinct sets of points (blue and red) originating the same 2D cumulative LitS.}
  \label{fig:potential-solutions}
\end{minipage}
\end{figure}

Since LitS convey detailed spatial information about $Q_\lambda$, it is natural to ask whether their construction from $Q_\lambda$ is a lossless process. If so, we could invert it to retrieve $Q_\lambda$ from them, making LitS a \emph{neighborhood transform}. This is an interesting trait because it allows us to look at neighborhood configurations from two perspectives. As they do not keep track of multiplicities, it is easy to see that regular LitS are not a neighborhood transform. Clearly, cumulative LitS cannot be either if $\varphi=0$ or $\varphi\geq\pi$. Nevertheless, a look at Figure~\ref{fig:cumulative-lits-3d} might convince us that cumulative LitS are a neighborhood transform for $\varphi\in(0,\pi)$.

Let us start with 3D cumulative LitS. To prove they are a neighborhood transform, it will suffice to show that it is always possible to locate a single illuminating neighbor $q\in Q_\lambda$ from a given cumulative LitS $L_p^+[\lambda,\varphi]$. Indeed, then we can subtract $c_\varphi(q)$ from $L_p^+[\lambda,\varphi]$ and iterate until we locate all of them and are left with the $0$ function on $S^2$. Since $Q_\lambda$ is finite and $L_p^+[\lambda,\varphi]$ was built according to Subsection~\ref{subsec:3d-lit-angles}, we know that every discontinuity circular arc found in $L_p^+[\lambda,\varphi]$ will be part of (and determines) the boundary of $c_\varphi(q)$ for a certain unknown $q\in Q_\lambda$. However, knowing $c_\varphi(q)$ tells us that $q$ lies on the ray from $p$ through $c_\varphi(q)$'s center. The defining condition of \eqref{eq:c-of-q}, which must be met with equality at $c_\varphi(q)$'s boundary, allows us to determine $q$ within that ray.

The argument above is not valid for 2D cumulative LitS because the discontinuity set of cumulative LitS in the plane consists of single points rather than circular arcs. The most we can say based on a single point discontinuity is that an illuminating neighbor lies on a ray that starts at the discontinuity point and makes an angle of $\varphi$ with respect to the normal direction. Figure~\ref{fig:potential-solutions} illustrates a scenario where $L_p^+[\lambda,\varphi]$ has two positive discontinuities $\alpha_0$ and $\alpha_1$, and two negative discontinuities $\beta_0$ and $\beta_1$, with respect to counterclockwise orientation. The two blue points are a valid solution for $Q_\lambda$, but so are the two points in red, allowing us to conclude that 2D cumulative LitS are not a neighborhood transform.

\section{Using LitS}\label{sec:using-lits}
While 3D point clouds, in actuality, come primarily from two main technologies—LiDAR and photogrammetry \cite{photogrammetry}—, their analysis and use in general permeate through a wide range of research areas and technological fields \cite{pc_applications_2021, pc_applications_2025}. Due to its high abstraction nature, LitS is best suited as an aiding tool in point cloud analysis, working as a piece of a larger engine rather than as an application in and of itself.

We devote Subsection~\ref{subsec:boundary-detection} to spelling out how LitS could be used for \emph{boundary detection}. Then we study LitS behavior for corner- and line-type neighborhoods, analyzing the dependency between underlying quantities like corner aperture and length of LitS' support. Finally, we move on to providing some visual representations of LitS properties to give an insight into other potential applications. Since LitS is a high-dimensional object, the drawings shown in this section are of properties of LitS, such as their maximum value or their total variation. They cannot possibly paint the whole picture, but they still offer a valuable peek into the behavior and traits of LitS in various situations.

\subsection{Interior and Boundary points}\label{subsec:boundary-detection}
\begin{figure}[h!]
\centering
\includegraphics[width=\linewidth]{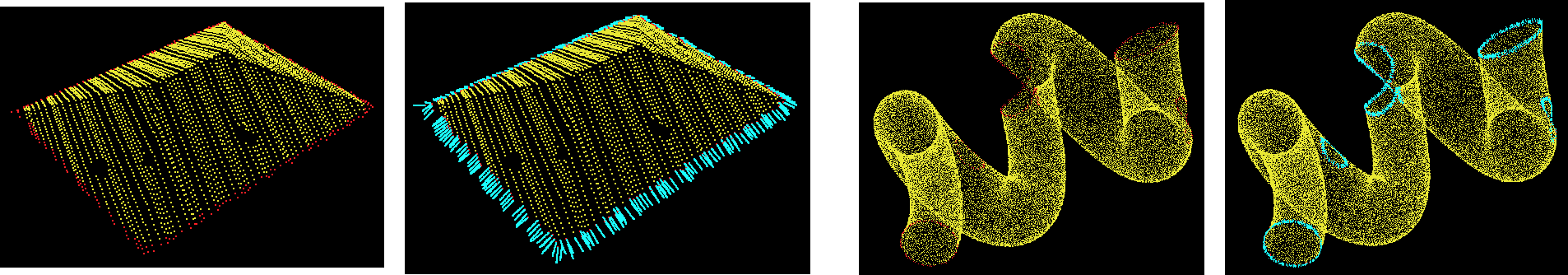}
\caption{Segmented roof (left) and simulated multiple-holed cellentani pasta (right), with boundary points in red and outside directions in light blue.}
\label{fig:roof-cellerani}
\end{figure}

Based on the notion of surroundedness introduced in Subsection~\ref{subsec:surroundedness}, and having chosen a sensible value of $\varphi$, we could stipulate that $p\in P$ is an \emph{interior point} if $p$ is $\varphi$-surrounded, and a \emph{boundary point} otherwise. Furthermore, if $p$ is a boundary point, then $L_p[\lambda,\varphi]$ will not be identically $1$, and so there must be a maximal interval in $S^1$ where $L_p[\lambda,\varphi]$ evaluates to $0$, the center of which should be a good approximation of an outside direction, as defined in Subsection~\ref{subsec:surroundedness}. See Figure~\ref{fig:roof-cellerani}. Moreover, the center of a maximal interval of $S^1$ where $L_p[\lambda,\varphi]$ evaluates to $1$ gives away an inside direction.

\begin{figure}[h!]
\centering
\includegraphics[width=.9\linewidth]{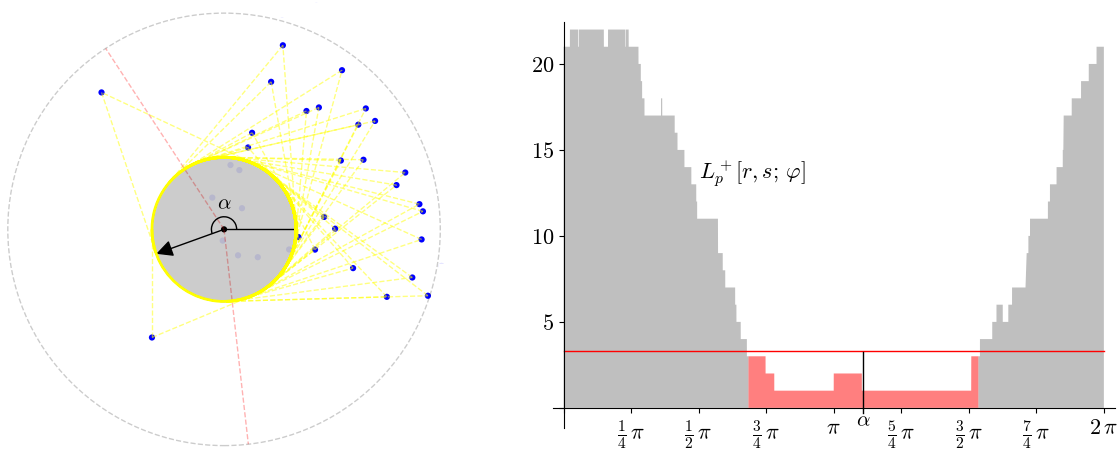}
\caption{Neighborhood with two noise points (left) and corresponding graph of $L_p^+[\lambda,\varphi]$ (right). A threshold $i_0$, set to $15\%$ of the maximum value of $L_p^+[\lambda,\varphi]$, determines a unique interval whose center $\alpha$ points to an outside direction.}
\label{fig:cumulative-lit-angles-boundary}
\end{figure}

However, real-world point clouds are often not short of outliers, which regular LitS do not resist. As shown in Figure~\ref{fig:cumulative-lit-angles-boundary}, a moderate amount of noise might cause boundary points to be misclassified. To get around this issue, employ cumulative LitS instead, choosing a threshold $i_0$ and stipulating that $p\in P$ is an interior point if $p$ is $\varphi$-surrounded with multiplicity $i_0$, and a boundary point otherwise. Now, if $p$ is a boundary point, then outside and inside directions can be read from maximal intervals of $S^1$ where $L_p^+[\lambda,\varphi]<i_0$ and $L_p^+[\lambda,\varphi]\geq i_0$, respectively.

Notice that, since neighbors at a distance lower than $r_p$ are ignored during LitS' construction, points that are closer than $r_p$ to a boundary will likely be classified as boundary points. In consequence, detected boundaries and edges will have a thickness modulated by $r_p$.

\subsection{Corner- and Line-Type Neighborhoods}
A good deal of LitS' appeal as a geometric descriptor is the fact that different neighborhood types generate LitS with a distinctive shape. In this subsection we will have a closer look at its behavior for corner-like and line-like neighborhoods. Since analyzing each type will require us to keep note of two different quantities already, we will assume throughout this subsection the standard limiting angle of incidence $\varphi=\pi/2$ for clarity's sake.

Let $Q$ be a neighborhood centered at a point $p$. Roughly speaking, we say $Q$ is a \emph{$\lambda$-corner} with aperture $\alpha$ and direction $\theta$ if the range of angular coordinates of points in $Q_\lambda$ is equal to $\alpha$, and $Q_\lambda$ is uniformly distributed on the annular region $r_p\leq r\leq r_Q$, $\theta-\alpha/2\leq t \leq\theta +\alpha/2$. Notice that parameterizing this definition by $\lambda$ allows us to disregard neighbors that are too close to $p$, whose angular coordinates might enlarge the neighborhood's angular range to a value that's not representative of its shape. For instance, disregarding its two leftmost outliers, the neighborhood in Figure~\ref{fig:cumulative-lit-angles-boundary} is a $1/3$-corner with aperture close to $\pi/2$, but the angular range of inlier points in $Q$ exceeds $\pi$.

\begin{wrapfigure}{l}{0.36\textwidth}
\includegraphics[width=\linewidth]{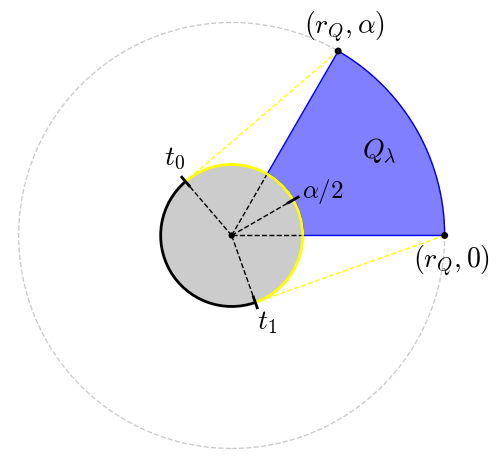}
\caption{Illustration of the annular sector where neighbors in $Q_\lambda$ must be contained for a $\lambda$-corner with aperture $\alpha$. The maximal arc possibly lit up by $Q_\lambda$ is determined by its outer corners, which in turn determines a lower bound for the length of LitS' zero set.}
\label{fig:L0}
\end{wrapfigure}
Since LitS are rotationally invariant, the particular direction $\theta$ the neighborhood happens to be centered at will not be relevant. The two quantities we will be interested in that actually shape the neighborhood are $\lambda$ and corner aperture $\alpha$. To understand the interplay between these two and the resulting LitS, we might start by considering the example in Figure~\ref{fig:cumulative-lit-angles-boundary}, and observing that a corner neighborhood will always yield a LitS function consisting of a single hill and valley. Furthermore, in the absence of outliers, the \emph{zero set} of LitS will be exactly one angular interval of $S^1$. Clearly this interval will vary with each neighbor configuration of $Q_\lambda$. However, if $Q$ is a $\lambda$-corner with aperture $\alpha$, then illuminating neighbors will be confined to an annular sector with angular range $\alpha$ between radii $r_p$ and $r_Q$. These constraints are enough to work out a theoretical lower bound for its length: As depicted in Figure~\ref{fig:L0}, assume $\alpha/2$ to be direction of $Q_\lambda$, without loss of generality. As further neighbors illuminate wider arcs, the maximal arc that can possibly be lit up by $Q_\lambda$ will be determined by the two points with polar coordinates $(r_Q,\alpha)$ and $(r_Q, 0)$. Using (\ref{eq:a_of_q}) we can determine the angular ends $t_0$ and $t_1$ of the $S^1$-complement of that maximal arc; $t_0=\alpha+\arccos(r_p/r_Q)$ and $t_1=2\pi-\arccos(r_p/r_Q)$. Taking into account that $r_p/r_Q=\lambda$ it follows that
\begin{equation}\label{eq:L0}
|t_1-t_0|=2\pi-\alpha-2\arccos\lambda.
\end{equation}
As a consequence, the zero set of LitS for a $\lambda$-corner neighborhood $Q$ with aperture $\alpha$ will have a length of at least this quantity.

Other than the size of its zero interval, there are a number of different quantities we can measure from the LitS of a corner-type neighborhood that encode geometric information about it. For instance, taking into account the representation of LitS derived in Subsection~\ref{subsec:visible-regions}, one might hope to measure density variations by looking at LitS' differences between close values. Or, as Figure~\ref{fig:dragon} demonstrates, even use LitS' \emph{total variation} as a metric of global angular uniformity of the neighborhood. Extracting a few selected measurements like these, it might be possible to characterize certain types of neighborhoods, such as corners. If the characterization has little overlap with other types of neighborhoods, LitS will be useful in detecting them within a scene. Figure~\ref{fig:lidar0cut-piece} should serve as indication of LitS' potential for pattern recognition tasks, particularly for corner detection from a selection of LitS measurements.

\begin{figure}[h!]
\centering
\includegraphics[width=\linewidth]{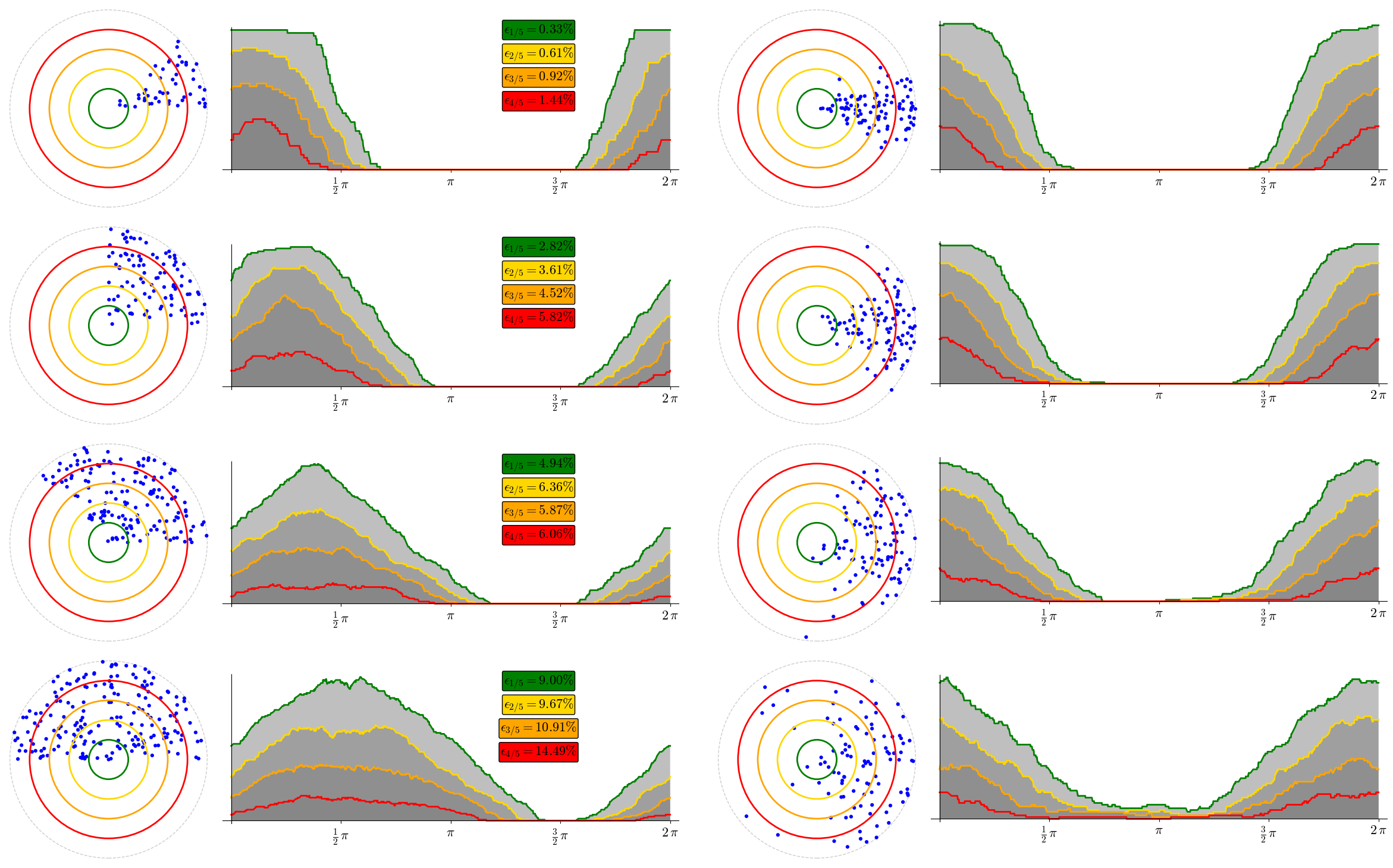}
\caption{Eight corner-type neighborhoods, each with associated LitS graphs for different $\lambda$ values. On the left, we used a uniform point distribution and apertures $\pi/4$, $\pi/2$, $3\pi/4$ and $\pi$ to model corners. For contrast, on the right we chose \emph{von Mises} distributions with common location parameter $\mu=0$, and concentration parameters $\kappa=16,\ 8,\ 4,\ 2.$ LitS with different $\lambda$ parameters were plotted next to each neighborhood, with heights normalized to the highest LitS value. $\lambda$ values from $0.2$ to $0.8$ in equally-spaced intervals of $0.2$, color-coded in green, yellow, orange and red, respectively. Colored boxes on the left also show the relative error $\epsilon_\lambda$ made by estimating the actual length of the corresponding zero interval using the lower bound provided by (\ref{eq:L0}).}
\label{fig:corners}
\end{figure}

We represent some relevant use cases in Figure~\ref{fig:corners} to show how LitS-derived metrics can be used for corner identification. Observe how all LitS shown do have the same shape, with one single peak and valley, and how the maximum absolute slope correlates with corner aperture. And, regarding the relative errors with respect to the lower bound in (\ref{eq:L0}), that two trends emerge: First, wider corner apertures leads to smaller zero intervals, naturally amplifying relative errors. Second, as $\lambda$ increases, so does the relative error. This is because the distance to the outer corners of the annular sector, measured from the closest neighbor, becomes more significant relative to $r_p$, which grows linearly with $\lambda$.

Let's now turn our attention to line-type neighborhoods. We could define a neighborhood $Q$ around point $p$ to be \emph{a line} with width $w$ and axial direction $d$ if points in $Q$ lie within distance $w$ of a line through $p$ in direction $d$, and are uniformly distributed there. Starting from this definition we could perform an analysis similar to the corner case above. It is easy to see that a line neighborhood will always consist of two peaks, corresponding to the underlying line directions; and two valleys, differing $\pi/2$ radians from the peaks. And, even though it is not so straightforward in this case, it is not hard to see that the four outer corners of the line region will illuminate $p$'s LitS maximally among all possible line neighborhoods with the same width and axial direction. So we could also analyze the size of the two zero intervals that appear for line neighborhoods. However, the resulting lower bound will not behave as well as (\ref{eq:L0}), since, for values of $w$ greater than $r_p$, the underlying line region is no longer split in two by $B(p,r_p)$, and so $p$'s LitS could be completely illuminated.

\begin{figure}[h!]
\centering
\includegraphics[width=\linewidth]{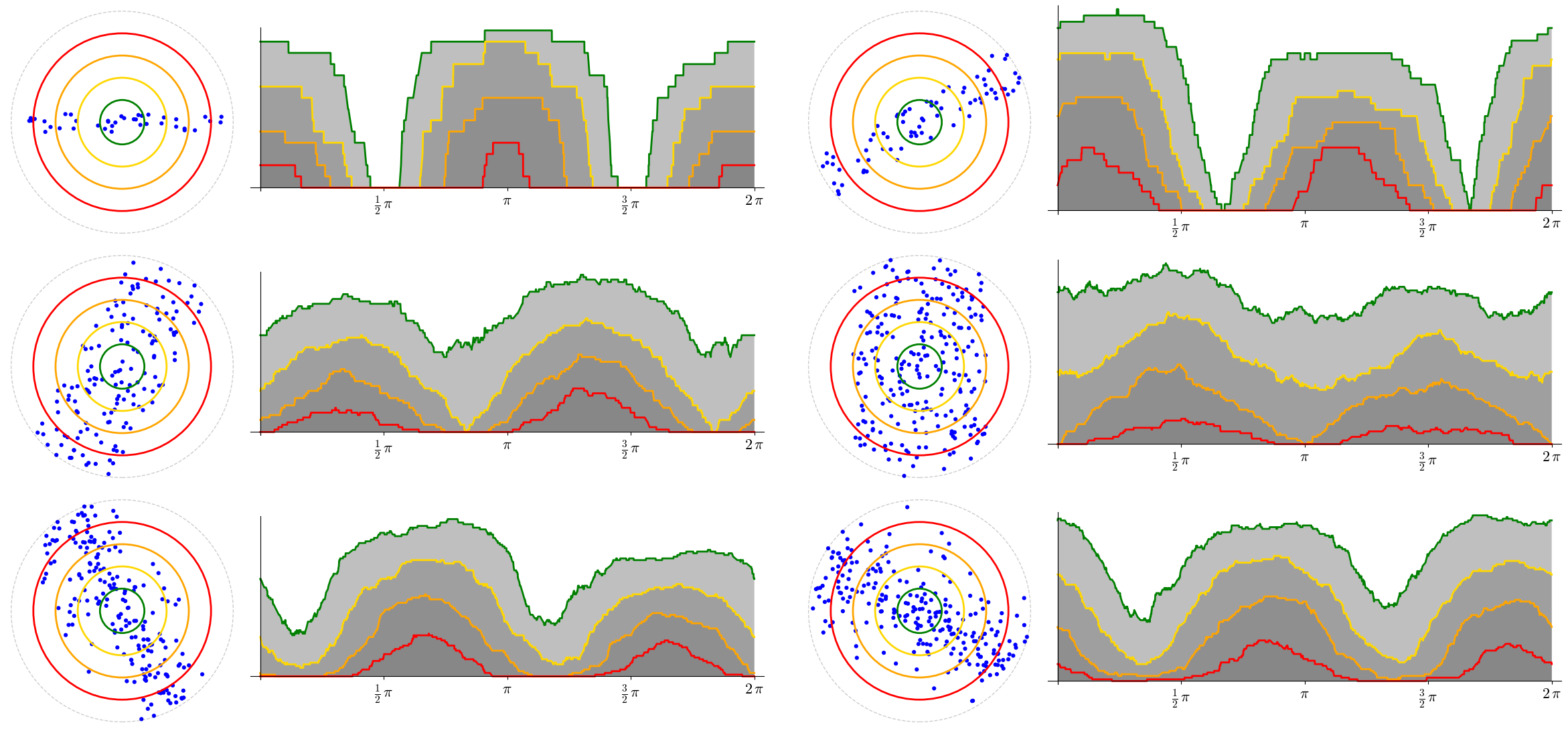}
\caption{Six line neighborhoods with orientations $i\cdot \pi/6$ for $i=0,\ldots,5$. LitS with $\lambda=0.2,\,0.4,\ 0.6$ and $0.8$ were plotted next to each neighborhood, with normalized heights and corresponding to colors green, yellow, orange and red, respectively. The first two rows display uniform point distributions and widths $10\%$, $20\%$, $40\%$ and $60\%$ of $r_Q$. In the last row we chose widths of $60\%$ and $80\%$ of $r_Q$ but changed to a triangular distribution in the line's perpendicular direction. This causes a relative decrease in the number of points near the edges of the line, which would happen with the spacing between scanlines as the distance increases from a TLS scanner placed at the center of the circles.}
\label{fig:lines}
\end{figure}

Again, there are many metrics that we can extract from LitS that help characterize line neighborhoods. Figure~\ref{fig:lines} shows a wide range of scenarios to give a comprehensive understanding of LitS behavior for line neighborhoods.

\subsection{Visualizing LitS}\label{subsec:visual}
There is a number of choices to analyze about a point's LitS in a 3D point cloud, starting at whether we want to compute regular or cumulative LitS. All neighborhoods considered in this section will be spherical, so we must specify a neighborhood radius. Notice, though, that this radius will be effectively variable in those figures where the illuminating neighborhood consists of the "\emph{k Nearest Neighbors (KNN)}", for a chosen value of $k$. Then a radius for $p$ or, equivalently, a value for $\lambda$ must be chosen. Since 3D is the most exciting case, all drawings are of 3D point clouds, which means we also have to choose a normal vector. In most drawings, we use the normal vector of the point's tangent plane because it is the most natural choice. However, other options might also be appropriate for specific applications and types of point clouds. Finally, we have to choose a limiting angle of incidence. While $\pi/2$ is the most straightforward choice, higher values may be more suitable for low-density point clouds, and lower values for high-density ones could be more sensible.

The drawings below display a hand-picked selection of some scalar properties of LitS. Even though they cannot possibly represent all of LitS’ possibilities, they should provide the reader with enough information to assess their behavior and usefulness in any given problem.
\begin{figure}[h!]
\centering
\includegraphics[width=\linewidth]{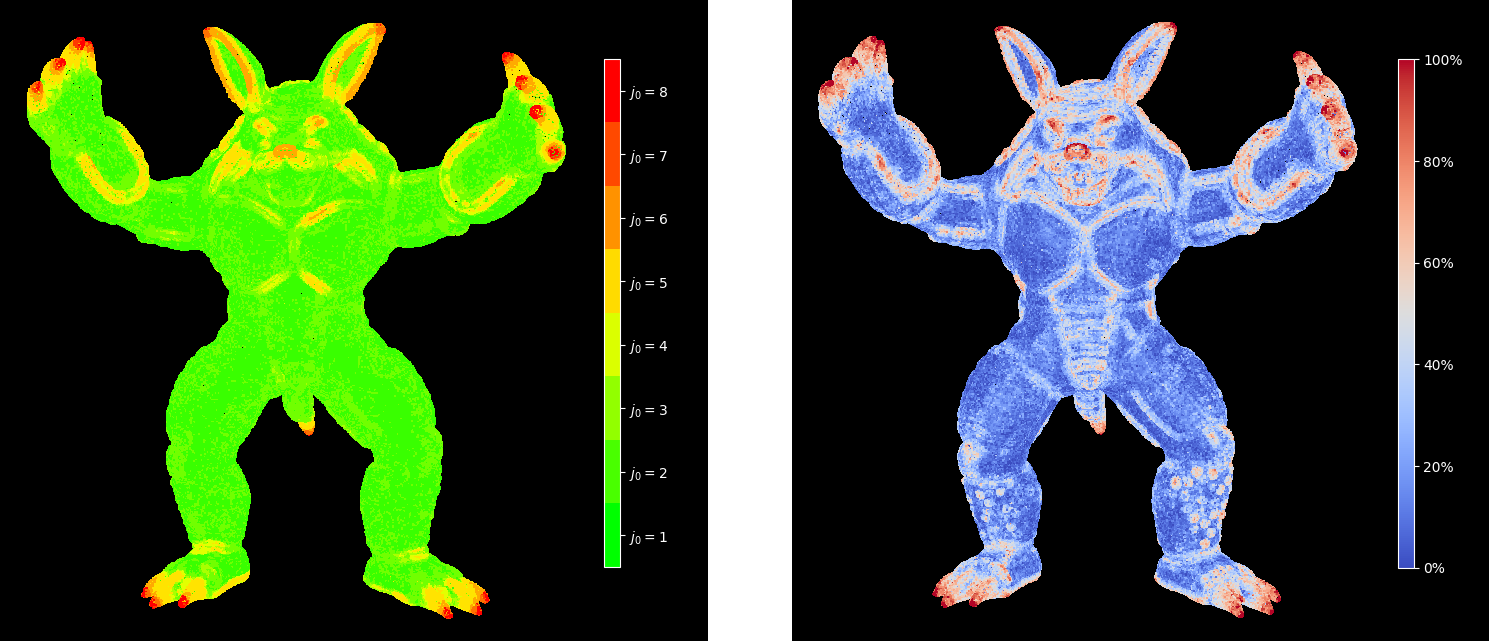}
\caption{Two LitS drawings of model "Stanford Armadillo" (\href{http://graphics.stanford.edu/data/3Dscanrep/}{graphics.stanford.edu/data/3Dscanrep/}).}
\label{fig:armadillo}
\end{figure}

\begin{wrapfigure}{r}{0.4\textwidth}
\includegraphics[width=0.9\linewidth]{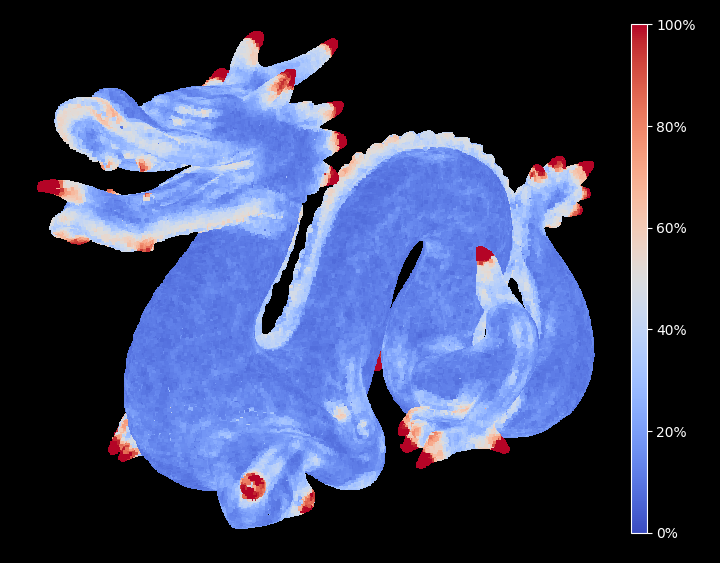}
\caption{Cumulative LitS drawing of model "Stanford Dragon".\vspace{-0.8cm}}
\label{fig:dragon}
\end{wrapfigure}
In Figure~\ref{fig:armadillo}, two different LitS drawings of the same model are shown. Both are regular LitS with $\lambda=2/3$, and both are computed with respect to the usual normal vector. In the left one, where the underlying neighborhoods have a common radius of 4.2, we chose a \emph{base incidence angle} $\varphi_0:=\pi/8$ and compute each point's \emph{surroundedness class} as the first integer $j_p$ for which $L_p[2/3,j_p\cdot\varphi_0](t)>0,\ t\in S^1$. Then we drew points according to their surroundedness class, with lower values in green to higher values in red. Observe this corresponds to a discretization of $\varphi^\ast$ (see Subsection~\ref{subsec:surroundedness}). On the right side, we used 20NN neihgborhoods and the canonical $\pi/2$ as the limiting angle of incidence, and a \emph{coolwarm} scale to indicate the proportion of LitS's domain that is unlit (i.e., that has a Lit value of 0). Notice that green-to-red color codes correlate with the degree of singularity of each point's neighborhood in the left drawing, and how using just the 20 nearest neighbors affects the level of detail in the right one.

Figure~\ref{fig:dragon} shows a drawing from cumulative LitS. We compute LitS along the normal vector, using a $\lambda$ value of $2/3$, and limiting angle of incidence of $3\pi/4$. Here, each point's percentile value of LitS total variation is represented using a coolwarm scale. Thus, a point in white, which is exactly in the middle of the scale, means 50\% of points yield LitS with higher total variation. We applied a \emph{moving average} beforehand to lower the sensitivity to high-frequency jumps. Notice that points with a high red intensity occur in regions of the model that differ greatly from being flat.

\begin{figure}[h]
\centering
\includegraphics[width=\linewidth]{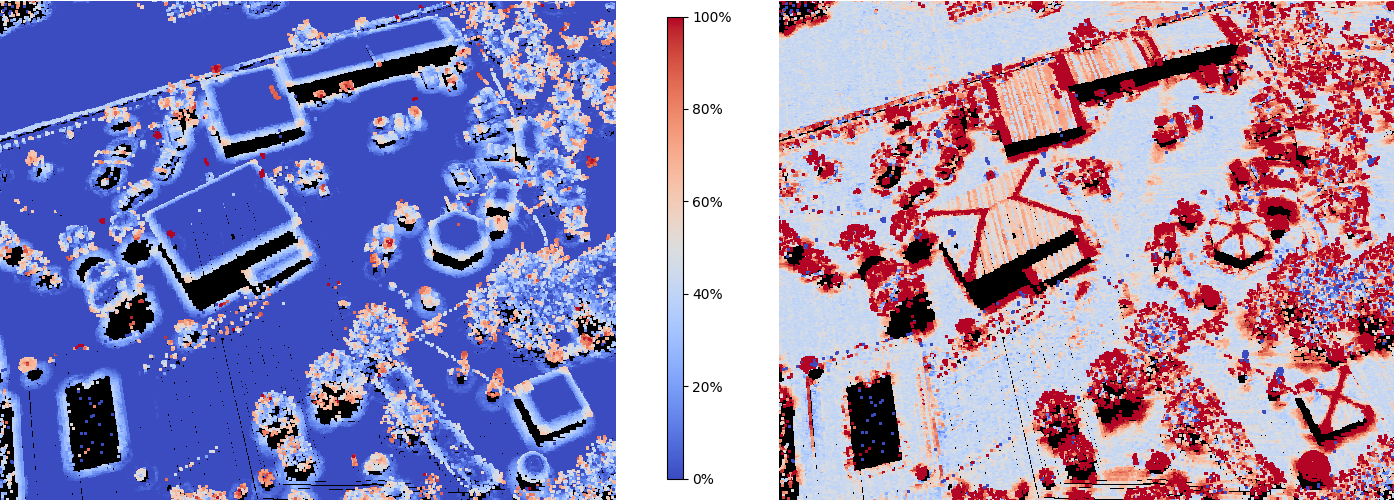}
\caption{Two LitS drawings of an airborne LiDAR point cloud from a rural region in Galicia.}
\label{fig:lidar0cut-piece}
\end{figure}

As an example of using LitS on airborne LiDAR point clouds, figure~\ref{fig:lidar0cut-piece} shows two drawings of the same region. Both are of cumulative LitS with $r_p=0.8$ and $r_Q=1.2$, and use a coolwarm colorscale. In the left one, the usual normal was chosen along with a limiting angle of incidence of $\pi/3$. Similarly to what we did in the right drawing of Figure~\ref{fig:armadillo}, here the scale draws the proportion of LitS domain that maps below a threshold of 15\% of the number of illuminating neighbors. On the right side, instead of the usual normal we chose a vector on its orthogonal complement with maximal $Z$-component, and a limiting angle of incidence of $3\pi/4$. Here the scale is proportional to the range of points' LitS, after applying a log-transformation to improve contrast. Notice the correlation between local maxima and corner points on the left and how the choice of normal vector on the right has caused roof joints to be warmer while roof edges with a constant height got relatively cooler.

\section{Conclusion and Future Work}
In this work, we introduced LitS, a novel neighborhood descriptor for point clouds. We provided a precise definition, an analysis of its main properties, and a graphical overview of their use in different situations. Even though only 3D cumulative LitS are a neighborhood transform, LitS, in general, is a rich and versatile neighborhood descriptor, which could come in handy whenever geometric factors are at play.

In an effort to keep its size moderate, a number of assumptions were made throughout this article. For example, we assume a constant limiting angle of incidence $\varphi$ for all illuminating neighbors, and we make a crude illuminance estimate by assuming that illuminating neighbors light up angle intervals uniformly. Also, we focused on 2D and 3D point clouds, but the ideas behind LitS' definition make sense regardless of the dimension of the ambient space. Removing these simplifying assumptions could open up promising directions for further inquiry.

In addition, there are research avenues that were not pursued here because they exceed the intended scope of this study. One such direction concerns the search for ways to reduce the size of cumulative LitS. As a neighborhood transform, they retain every single bit of information about a point's illuminating neighborhood, so it could be useful to come up with lighter approximations, especially for large neighborhoods and dense point clouds. Another direction follows from the realization that every measurement that we might perform on LitS will be a point cloud descriptor in its own right. Figures~\ref{fig:armadillo} to \ref{fig:lidar0cut-piece} display some of them, but there could be value in a more systematic and comprehensive exploration of such descriptors.

Furthermore, it could be interesting to explore possible use cases of LitS as a research tool in several areas. For instance, LitS could have potential in \emph{classification and segmentation tasks}, both as an intermediate step in segmentation and as a presegmentation. Alternatively, since LitS are essentially a way to intuitively encode (most notably angular) information about a point's neighborhood, it is reasonable to expect them to be a valuable tool in \emph{direction-based outlier detection}. Or, considering all corner and edge points present a similar cumulative LitS graph with a single peak and valley (see Figure~\ref{fig:corners}), one could use the size of the maximal below-some-threshold interval around the valley's abscissa to \emph{identify corner point candidates}. Actually, given that, often, points with an identifiable neighborhood distribution also have a visually distinctive LitS graph, analyzing how LitS change between close points could be an effective way to tackle \emph{pattern detection} problems. Lastly, every measurable property of LitS could be a relevant \emph{input feature for machine learning models}. In particular, as almost every cumulative LitS have that sum-of-low-frequency-sinusoids look, they should be well approximated with just a handful of terms of its Fourier series expansion, which could be an interesting input feature.

\section*{Acknowledgements}
This work has received financial support from the Agencia Estatal de Investigación (Spain) (PID2022-141623NB-I00), the Xunta de Galicia - Consellería de Cultura, Educación, Formación Profesional e Universidades (Centro de investigación de Galicia accreditation 2024-2027 ED431G-2023/04 and Reference Competitive Group accreditation ED431C-2022/016) and the European Union (European Regional Development Fund - ERDF).

\bibliography{references}

\end{document}